\documentclass[10pt]{article} % For LaTeX2e
\usepackage[preprint]{tmlr}
\usepackage{hyperref}
\usepackage{url}

\usepackage{microtype}
\usepackage{graphicx}
\usepackage{subfigure}
\usepackage{booktabs} 

\usepackage{amsmath}
\usepackage{amssymb}
\usepackage{mathtools}
\usepackage{amsthm}
\usepackage{bbm}

\usepackage[capitalize,noabbrev]{cleveref}

\usepackage[toc,page,header]{appendix}
\usepackage{minitoc}

\theoremstyle{plain}
\newtheorem{theorem}{Theorem}[section]
\newtheorem{proposition}[theorem]{Proposition}

\theoremstyle{definition}
\newtheorem{definition}[theorem]{Definition}

\theoremstyle{remark}

\usepackage[toc,page,header]{appendix}
\usepackage{minitoc}
\usepackage{hyperref}       % hyperlinks
\usepackage{url}            % simple URL typesetting
\usepackage{booktabs}       % professional-quality tables
\usepackage{amsfonts}       % blackboard math symbols
\usepackage{nicefrac}       % compact symbols for 1/2, etc.
\usepackage{microtype}      % microtypography
\usepackage{xcolor}     
 
\usepackage{mathtools}
\usepackage{amssymb}
\usepackage{amsthm}
\usepackage{amsfonts}
\usepackage{bm}
\usepackage{bbm}

\usepackage{amsmath}

\usepackage{adjustbox}
\usepackage{multirow}
\usepackage{subfigure}
\usepackage{wrapfig}
\usepackage{placeins}
\usepackage{float}

\usepackage{algorithm}

\usepackage{algorithmic}
\usepackage{amsmath}
\usepackage{mathtools}
\usepackage{dsfont}
\usepackage{caption,subfigure,graphicx,epstopdf,multirow,multicol,booktabs,verbatim,wrapfig,bm}
\usepackage{makecell}
\usepackage{amsthm,amssymb,amsfonts,amsmath}
\usepackage{algorithmic}
\usepackage{subfiles}

\title{Interpolation for Robust Learning: \\
Data Augmentation on Wasserstein Geodesics}

\def\month{MM}  % Insert correct month for camera-ready version
\def\year{YYYY} % Insert correct year for camera-ready version
\def\openreview{\url{https://openreview.net/forum?id=XXXX}} % Insert correct link to OpenReview for camera-ready version

\begin{document}

\maketitle

\vspace*{-2cm}

%\centering
\begin{center}
{\large 
\begin{tabular}{cccc}
Jiacheng Zhu$^{1}$ &  Jielin Qiu$^{1}$ & Aritra Guha$^{2}$ & Zhuolin Yang$^3$ \\
\end{tabular}

\begin{tabular}{ccc}
      XuanLong Nguyen$^4$ & Bo Li$^3$ & Ding Zhao$^{1}$ 
\end{tabular}
}

% \vspace*{.2in}

{
\begin{tabular}{c}
Carnegie Mellon University$^1$, AT\&T Chief Data Office$^2$ \\
University of Illinois at Urbana-Champaign$^3$, University of Michigan$^4$ \\
\end{tabular}
}
\vspace*{.2in}
\end{center}

\begin{abstract}
We propose to study and promote the robustness of a model as per its performance through the interpolation of training data distributions. 
Specifically, (1) we augment the data by finding the worst-case Wasserstein barycenter on the geodesic connecting subpopulation distributions of different categories. 
(2) We regularize the model for smoother performance on the continuous geodesic path connecting subpopulation distributions.
(3) Additionally, we provide a theoretical guarantee of robustness improvement and investigate how the geodesic location and the sample size contribute, respectively. 
Experimental validations of the proposed strategy on \textit{four} datasets, including CIFAR-100 and ImageNet, establish the efficacy of our method, e.g., our method improves the baselines' certifiable robustness on CIFAR10 up to $7.7\%$, with $16.8\%$ on empirical robustness on CIFAR-100.
Our work provides a new perspective of model robustness through the lens of Wasserstein geodesic-based interpolation with a practical off-the-shelf strategy that can be combined with existing robust training methods. 
\end{abstract}

\doparttoc % Tell to minitoc to generate a toc for the parts
\faketableofcontents

\section{Introduction}
Deep neural networks (DNNs) have shown tremendous success in an increasing range of domains such as natural language processing, image classification \& generation and even scientific discovery (e.g.~\citep{bahdanau2014neural_nlp,krizhevsky2017imagenet,ramesh2022hierarchical_clip_dalle2}).
However, despite their super-human accuracy on training datasets, neural networks may not be robust. 
For example, adding imperceptible perturbations,  e.g., \textit{adversarial attacks}, to the inputs may cause neural networks to make catastrophic mistakes~\citep{Szegedy2014IntriguingPO,goodfellow2014explaining}. 
Conventional defense techniques focus on obtaining adversarial perturbations~\citep{carlini2017towards,goodfellow2014explaining} and augmenting them to the training process. 
For instance, the projected gradient descent (PGD)~\citep{madry2018towards}, which seeks the worst-case perturbation via iterative updates, marks a category of effective defense methods.

Recent works show that additional training data, including data augmentation and unlabeled datasets, can effectively improve the robustness of deep learning models~\citep{volpi2018generalizing_data_aug_dro,rebuffi2021data_aug_rob_deepmind}. 
Despite augmenting worst-case samples with gradient information, an unlabeled, even out-of-distribution dataset may also be helpful in this regard~\citep{carmon2019unlabeled_imp_rob,bhagoji2019lower_adv_ot}. Among available strategies, Mixup~\citep{zhang2018mixup}, which interpolates training samples via convex combinations, is shown to improve both the robustness and generalization ~\citep{zhang2021mix_rob_gen}.
Moreover, Gaussian noise is appended to training samples to achieve certifiable robustness~\citep{cohen2019certified_cert_rob}, which guarantees the model performance under a certain degree of perturbation. 
However, most aforementioned approaches operate on individual data samples (or pairs) and require the specification of a family of augmentation policies, model architecture, and additional datasets. Thus, the robustness can hardly be generalizable, e.g., to out-of-sample data examples. 
%in terms of out-of-sample data examples, out-of-distribution domains, and various robustness measurements. 
To this end, we are curious: \textit{Can we let the underlying data distribution guide us to robustness?}
 
% Perspective from probability distributions
We propose a framework to augment synthetic data samples by interpolating the underlying distributions of the training datasets. 
Specifically, we find the \textit{worst-case interpolation distributions} that lie on the decision boundary and improve the model's smoothness with samples from these distributions. This is justified by the fact that better decision margin~\citep{kim2021bridged_margin} and boundary thickness~\citep{moosavi2019robustness} are shown to benefit robustness.

The probabilistic perspective allows us to formulate the distribution interpolation as barycenter distribution on \textit{Wasserstein geodesics}, where the latter can be reviewed as a dynamic formulation of optimal transport (OT)~\citep{villani2009optimal}. 
This formulation has the following benefits: 
(a). It nicely connects our worst-case barycenter distribution to distributionally robust optimization~\citep{DRO-duchi} on the geodesic and substantiates our method’s benefit in promoting robustness. 
(b). The geodesic provides a new protocol to assess one model's robustness beyond the local area surrounding individual data points.
(c). Our data augmentation strategy also exploits the local distribution structure by generalizing the interpolation from (mostly Euclidean) feature space to probability space.
(d). Apart from most OT use cases that rely on coupling, we reformulate the objective to include explicit optimal transport maps following McCann's interpolation~\citep{mccann1997convexity}. This enables us to deal with out-of-sample data and appreciate the recent advances in OT map estimation~\citep{perrot2016mapping,Zhu2021FunctionalOT,amos2022amortizing}, neural OT map estimators~\citep{makkuva2020optimal_icnn,seguy2017large,fan2021scalable_map} for large-scale high-dimensional datasets, and OT between labeled datasets~\citep{alvarez2020geometric_otdd,fan2022generating_geodesic}.

% Specifically, augmenting data with random Gaussian noise \citep{cohen2019certified_cert_rob} or transformations \citep{li2021tss_TSS} yields certifiable smoothed models. Mixup methods \citep{Zhang2018mixupBE,greenewald2021_kmixup}, which augment data with weighted averages of training points, also promote the certifiable robustness \citep{jeong2021smoothmix}. 

% Although it is natural to characterize the adversarial robustness in the local area around a sample, the concept of interpolation allows us to look for the space between samples where the decision boundary typically lies. In fact, \citep{zhang2018mixup} stated that Mixup smooths the classifier to reduce unnecessary oscillations between training examples. In addition, the decision margin~\citep{kim2021bridged_margin} and boundary thickness~\citep{wang2019improving,moosavi2019robustness} are also critical factors for robustness~\citep{wang2022removing_rob}.
% The interpolation of different data distributions traverses across the decision boundaries of the model, thus offering a new perspective to inspect the robustness of the predictive performance.

% {\color{blue}{AG: Introduce OT briefly and discuss why it may good to do mixup like strategy with OT}}

% {\color{blue}{AG: Don't make separate section for this. Put in Section 1 prior to our contributions. This puts our contributions in context}}

\subsection{Related Work}
\paragraph{Interpolation of data distributions}
~\cite{gao2021infor_geo_task} propose a transfer distance based on the interpolation of tasks to guide transfer learning. 
Meta-learning with task interpolation~\citep{yao2021meta_task_interp}, which mixes features and labels according to tasks, also effectively improves generalization. 
For gradual domain adaptation, ~\cite{wang2022understanding_gda} interpolate the source dataset towards the target in a probabilistic fashion following the OT schema. Moreover, the interpolation of environment distribution~\cite{huang2022curriculum} is also effective in reinforcement learning.
The recently proposed rectified flow~\citep{liu2022_rectified_flow} facilities a generating process by interpolating data distributions in their optimal paths. In addition, the distribution interpolant~\citep{albergo2022building_interpolant} has motivated an efficient flow model which avoids the costly ODE solver but minimizes a quadratic objective that controls the Wasserstein distance between the source and the target.
Recently, the interpolation among multiple datasets realized on generalized geodesics~\citep{fan2022generating_geodesic} is also shown to enhance generalizability for pretraining tasks. 
Most aforementioned works focus on the benefit of distribution interpolation in generalization. 

On the other hand, Mixup~\citep{zhang2018mixup}, which augments data by linear interpolating between two samples, is a simple yet effective data augmentation method. 
As detailed in survey~\citep{cao2022survey_mixup,lewy2022overview_survey_mixup}, there are plenty of extensions such as CutMix~\citep{yun2019cutmix}, saliency guided~\citep{uddin2020saliencymix}, AugMix~\citep{hendrycks2019_augmix}, manifold mixup~\citep{verma2019manifold}, and so on~\citep{yao2022_LISA}.
A few studies have explored the usage of optimal transport (OT) ideas within mixup when interpolating features~\citep{kim2020_puzzlemix,kim2021_co-mixup}. 
However, those methods focus on individual pairs, thus neglecting the local distribution structure of the data. One recent work~\citep{greenewald2021_kmixup} also explores mixing multiple-batch samples with OT alignment. 
Although their proposed K-mixup better preserves the data manifold, our approach aims to determine the worst-case Wasserstein barycenter, achieved through an interpolation realized by transport maps.

\paragraph{Data augmentation for robustness.}
%Despite the success of many heuristic defense methods~\citep{shorten2019survey_data_aug}, 
Augmentation of data\citep{rebuffi2021data_aug_rob_deepmind,volpi2018generalizing_data_aug_dro} or more training data \citep{carmon2019unlabeled_imp_rob,sehwag2021robust_aux_distribution} can improve the performance and robustness of deep learning models. However,~\citep{schmidt2018adversarially} show that 
sample complexity for robust learning may be prohibitively large when compared to standard learning. Moreover, with similar theoretical frameworks, recent papers ~\citep{deng2021improving_ood_data_rob_gaussian_mix,xing2022unlabeled_unlabel_minimax} further establish theoretical justifications to
characterize the benefit of additional samples for model robustness. However, additional data may not be available; in this work, we therefore use a data-dependent approach to generate additional data samples. 

%under supervised or self-supervised learning. 
% {\color{blue}{AG: brief one line summary of what these recent works say? otherwise the statement sounds like it is hanging in air}}

\subsection{Contributions}

Our key contributions are as follows. We propose a data augmentation strategy that improves the robustness of the label prediction task by finding worst-case data distributions on the interpolation within training distributions. This is realized through the notion of Wasserstein geodesics and optimal transport map, and further strengthened by connection to DRO and  regularization effect. Additionally, we also provide a theoretical guarantee of robustness improvement and investigate how the geodesic location and the sample size contribute, respectively. Experimental validations of the proposed strategy on \textit{four} datasets including CIFAR-100 and ImageNet establish the efficacy of our method.

% We consider different subpopulations (classes) as distinct distributions for classification problems. We propose to characterize the robustness behavior of the classifier on a set of distributions that interpolate two different subpopulation distributions. 
% {\color{blue}{AG: this is one way to do interpolation, not the only way, rephrase this }}
% With optimal transport (OT)~\citep{villani2009optimal}, we know such interpolation is the Wasserstein barycenters that form a Wasserstein geodesic. In addition, we can realize such interpolation by a Monge map estimated from samples. 
% (1) Following the minimax adversarial training scheme, we propose a data augmentation framework that finds the worst-case barycenter on the geodesic. {\color{blue}{AG: avoid mentioning geodesic if it has not been described/mentioned before, make it something simpler.}}{\color{red}}Moreover, to pursue the smoothness for robustness, we propose a scheme to regularize the change in the model's loss when moving along the geodesic.
% (2) Theoretically, we establish that our data augmentation scheme, supported by an optimal transport map estimated from data, leads to improved adversarial robustness. To guarantee the improvement, we illustrate the desired geodesic location with its relation to the sample size. 
% (3) We validate our study on two image classification benchmarks, MNIST and CIFAR-10, for both empirical $\ell_\infty$ robustness and certified $\ell_2$ robustness. Our method outperforms baselines under different configurations. 

% {\color{red} A paragraph to stand away from empirical works?}

\section{Preliminaries}\label{sec:prelim}
Consider a classification problem on the data set $S = \{ (x_1, y_1),...,(x_n, y_n)\}$, where $x_i \in \mathcal{X} \subseteq \mathbb{R}^d$ and $y_i \in \mathcal{Y} \subseteq \mathbb{R}^y$ are drawn i.i.d from a joint distribution $\mathcal{P}^{all}_{x,y}$. Having a loss criteria $l(\cdot, \cdot): \mathcal{Y} \times \mathcal{Y} \mapsto \mathbb{R} $, the task is to seek a prediction function $f_\theta: \mathbb{R}^d \mapsto \mathbb{R}^y$ that minimizes the standard loss $L(\theta) = \mathbb{E}_{x,y \sim \mathcal{P}^{all}_{x,y}}[l(f_\theta (x), y)]$, and in practice people optimize the empirical loss $\hat{L}(\theta) = 1/n \sum_{i=1}^n l(f(x_i), y_i)$ following the Empirical Risk Minimization (ERM). In large-scale classification tasks such as $k$-class image classification, the label is the one-hot encoding of the class as $y_i \in \{0, 1\}^k := \mathbb{R}^k$ and $l$ are typically cross-entropy loss.

%\subsection{Adversarial robustness}
\label{subsec:preliminary_on_robustness}
% \TBD{Less motivation}
\paragraph{Adversarial training \& distributional robustness } 
Typically, adversarial training is a minimax optimization problem~\citep{madry2018towards} which finds adversarial examples $x + \delta $ within a perturbation set $ \mathcal{S} = \{\delta: \| \delta \|_\infty \leq \epsilon, \epsilon > 0 \}$. While finding specific attack examples is effective, there are potential issues such as overfitting on the attack pattern~\citep{kurakin2016adversarial_pgd,xiao2019enhancing_adv,zhang2019defense_adv}.

An alternative approach is to capture the distribution of adversarial perturbations for more generalizable adversarial training~\citep{dong2020adversarial}.
In particular, the optimization problem solves a
distributional robust optimization \citep{DRO-duchi,weber2022certifying_cert_ood_gen_dro} as follows:
\begin{equation} %\small
    \min_{\theta} \sup_{Q_{x,y} \in \mathcal{U}_P} \mathbb{E}_{x, y \sim Q_{x,y}} [l(f_{\theta}(x), y)],
    \label{eq:eq1_dro}
\end{equation} 
where $\mathcal{U}_P \subseteq \mathcal{P(Z)} $ is a set of distributions with constrained support. Intuitively, it finds the worst-case optimal predictor $f^*_\theta$ when the data distribution $P$ is perturbed towards an adversarial distribution $\mathcal{U}_P$. 
However, the adversarial distribution family needs to be properly specified. In addition, adversarial training may still have unsmooth decision boundaries~\citep{moosavi2019robustness}, thus, are vulnerable around unobserved samples.

% {\color{blue}{AG: mention adversarial robustness may have unsmooth boundaries}}

% {\color{blue}{AG: shorten to one paragraph}}
\paragraph{Optimal transport}
Given two probability distributions $\mu, \nu \in \mathcal{M}(\mathcal{X})$, where  
$\mathcal{M}(\mathcal{X})$ is the set of Borel measures on $\mathcal{X}$. 
The optimal transport (OT)~\citep{villani2009optimal} finds the optimal joint distribution and quantifies the movement between $\mu$ and $\nu$. 
Particularly, the Wasserstein distance is formulated as 
$ W_p(\mu, \nu) := ( \inf_{\pi \in \Pi} \int_{\mathcal{X} \times \mathcal{X}} d^p(x, y) d \pi(x, y) )^{1/p},  $
% \begin{align} \small
%     W_p(\mu, \nu) := ( \inf_{\pi \in \Pi} \int_{\mathcal{X} \times \mathcal{X}} d^p(x, y) d \pi(x, y) )^{1/p}, 
% \end{align}
where $d(\cdot, \cdot) : \mathcal{X} \times \mathcal{X} \mapsto \mathbb{R}^+$ is the ground metric cost function, and $\Pi$ denotes set of all joint probability measures on $\mathcal{X} \times \mathcal{X}$ that have the corresponding marginals $\mu$ and $\nu$.
%The well-known Wasserstein distance originated from OT problem which aims at finding an optimal coupling $\pi$ that minimizes the transportation cost. 
%\textcolor{blue}{zhuolin: can we have a formal oral definition of Wassertein distance here?}
% \begin{definition}
% (Wasserstein Distances). For $p \in [1, \infty]$ and probability measures $\mu$ and $ \nu \in \mathcal{M}(\mathcal{X})$. The $p-$Wasserstein distance between them is defined as
% \begin{equation} \small
% \begin{aligned}
% W_p(\mu, \nu) := \left( \inf_{\pi \in \Pi} \int_{\mathcal{X} \times \mathcal{X}} d^p(x, y) d \pi(x, y) \right)^{1/p}, 
% %\text{  } (x, y) \in \mathcal{X} \times \mathcal{X}
% \label{eq:w-dist}
% \end{aligned}
% \end{equation}
% where $\Pi$ is the set of all probability measures on $\mathcal{X} \times \mathcal{X}$.
% \end{definition}
\paragraph{Wasserstein barycenter and geodesic}
Equipped with the Wasserstein distance, we can average and interpolate distributions beyond the Euclidean space.  The Wasserstein barycenter $\mu^{\{ \nu_i\}}_{\alpha}$ of a set of measures $\{\nu_1, ..., \nu_N \}$ in a probability space $\mathbb{P} \subset \mathcal{M}(\mathcal{X})$ is a minimizer of objective $U^N_{wb}$ over $\mathbb{P}$, where
\begin{align} \small
    U^N_{wb}(\mu) :=  \sum_{i=1}^N \alpha_i W(\mu, \nu_i),
    \label{eq:barycenter}
\end{align}
and $\alpha_i$ are the weights such that $\sum \alpha_i = 1$ and $\alpha_i > 0$. 
% The Monge formulation of optimal transport finds a map $T : \mathbb{R}^d \mapsto \mathbb{R}^d$ than transport a distribution $P$ towards $Q$:
% \begin{equation}
%   T^* = \arg \inf_T \int \|x - T(x) \|^p d P(x),
% \end{equation}
% where a minimizer $T^*$ is the optimal transport map such that $T^*_\# P = Q$, where $T^*_\# P$ is the pushforward of $P$. Given distributions $P$ and $Q$, if $T^*$ exits, then map $T_t(x) = (1 - t) x + t T^*(x)$ gives the path of a particle of mass at $x$ and $P_t = T_{t \#} P$ is the \textit{geodesic} connecting $P$ to $Q$. 
% The barycenter is the Fr\'echet mean, or the Wasserstein population mean~\citep{bigot2017geodesic_pca} with uniform weights.
Let $\nu_0$ and $\nu_1$ be two measure with an existing optimal transportation map, $T$, satisfying $\nu_1 = T\# \nu_0$ and $W^2_2(\nu_0, \nu_1) = \int_{\mathcal{X}} \|x -T(x)\|^2 d \nu_0$. Then, for each $t$, the barycenter distribution(denoted by $\mu_t$) corresponding to  $\alpha_0=t, \alpha_1=1-t$ in Eq.\eqref{eq:barycenter}, lies on the geodesic curve connecting  $\nu_0$ and $\nu_1$ . Moreover, 
%\AG{cite https://lchizat.github.io/files2020ot/lecture3.pdf}
~\citep{mccann1997convexity},
% \textit{McCann's interpolation} 
\begin{align}\small
    \mu_t:=((1-t)\text{Id} + t T)\# \nu_0 , 
    \label{eq:maccann}
\end{align}
% where $T = \nabla \phi$ transporting $\nu_0$ to $\nu_1$ is a Brenier's {\color{blue} push-forward} map $\nu_1 = T\# \nu_0$  
% The set of interpolation barycenters of $\nu_0$ and $\nu_1$ eq.(\eqref{eq:maccann}) define a constant speed geodesic $g(t), t\in [0, 1]$ connecting $g(0)=\nu_0$ and $g(1)=\nu_1$ such that for each barycenter $\mu_i$ \AG{what do you mean by each barycenter? }there exists a $t_i$ so that $\mu_i = g(t_i)$~\citep{agueh2011barycenters}.
where $\text{Id}$ is the identity map.
% It's worth mentioning that, as we have $\nu_1 = T\# \nu_0$, the above equation may reminds one of the convention mixup and we will later show mixup is one form of geodesic interpolation.
While the above McCann's interpolation is defined only for two distributions, the \textit{Wasserstein barycenter} eq. (\ref{eq:barycenter}) can be viewed as a generalization to $N \geq 2$ marginals. To this point, we are able to look for adversarial distributions leveraging such interpolation.

%%%%%%%%%%%%%%%%%%%%%%%%%%%%%%%%%%%%%%%%%%%%%%%%%%%%%
%%%%%%% Main section
%%%%%%%%%%%%%%%%%%%%%%%%%%%%%%%

\begin{figure*}[t!]
	\centering	
	\includegraphics[width=0.98\textwidth]{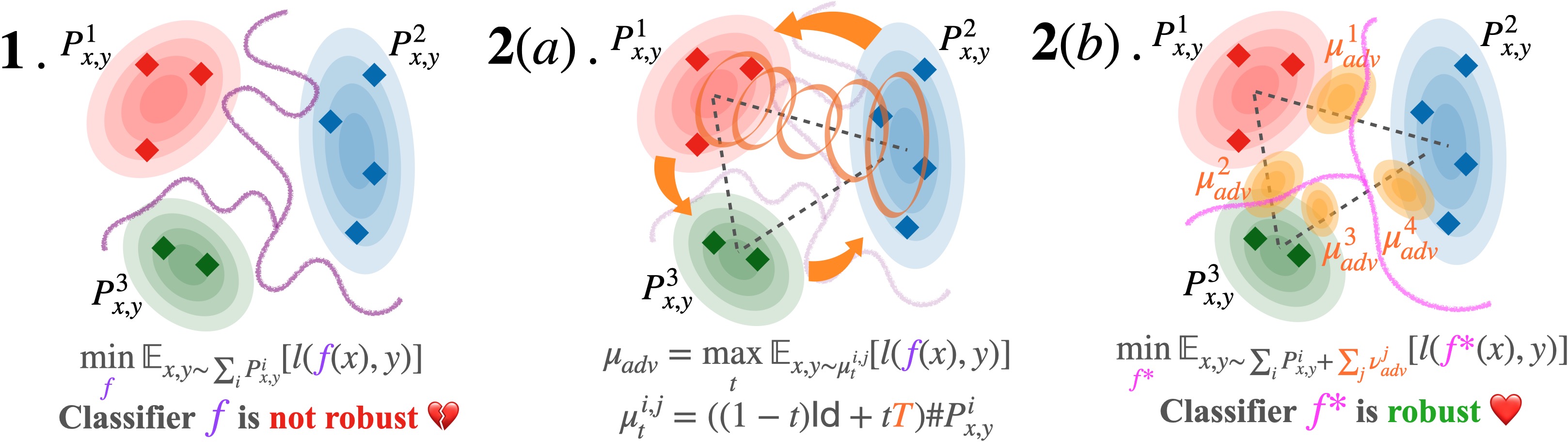}
	\vspace{-5pt}
	\caption{
	\textbf{1}: For a classification problem, the data samples give empirical access to the underlying data distribution, which is a mixture of distinct subpopulation distributions. \textbf{2(a)}: The inner maximization step finds the worst-case  Wasserstein barycenter on the geodesic. \textbf{2(b)}: The outer minimization step updates the predictive function with augmented samples.
	}
	\label{fig:fig1_illstration}
\end{figure*}

\section{Adversarial robustness by interpolation
%Dynamic formulation of robustness: \\ The adversarial Barycenter
}\label{sec:adv_interpolate}
For a $k$-classification task, it is natural to view the data of each category are samples from distinct distributions $P^{all}_{x,y}=\sum^k_i P^i_{x,y}$. 
% We restrict the adversarial distribution family to be the geodesic interpolations between individual subpopulation distributions. 
Following eq.(\ref{eq:eq1_dro}), we can write the following minimax objective 
\begin{align}
    \min_{f} \max_\alpha \mathbb{E}_{x, y \sim U^{wb}(\alpha) } \left[ l(f(x), y) \right] 
    % \label{eq:adv_w_bc}
\end{align}
where $U^{wb}(\alpha)$ is the Wasserstein barycenter (eq. \ref{eq:barycenter}), in other words, the interpolation of different category distributions. In fact, this corresponds to a special case of DRO eq.(\ref{eq:eq1_dro}) as shown in the below proposition, and naturally corresponds to the adversarial distribution within a particular Wasserstein ball, for distributional robustness.
% \subsection{DRO and Geodesic connection}
% The Distributionally robust optimization(DRO) setting~\citep{DRO-duchi} considers training the model on perturbed training data, which makes the model robust to distributions shifts. 
%Below we provide an example which relates the settings in this paper to the DRO setting.
\begin{proposition}
\label{prop:DRO-geodesic}
Suppose the original data $\{ {X}_i, {Y}_i, \}_{i=1}^{{n_0}}$ are iid with distribution satisfying $X|Y=y \sim \nu_{\mu}(X|Y)=N_d(y\mu, \sigma^2 I)$, $Y\sim \nu(Y)=U\{-1,1\}$. Assume the loss function is given by $l(f(x),y)=\mathbbm{1}(f(x)\neq y)$. We consider $f(x)=f_{\alpha,\beta}(x)$ of the form $\text{sign}(\alpha'x+\beta)$.Then the following holds
\begin{eqnarray}
\label{eq: maxinf-gaussian-geodesic}
& & \max_{A(\nu_{\mu},\epsilon)} \inf_{\alpha, \beta}  \mathbb{E}_{(X,Y) \sim \eta(X|Y)\nu(Y)}l(f_{(\alpha, \beta)}(X),Y)\nonumber\\
    &= & inf_{\alpha,\beta} \max_{A(\nu_{\mu},\epsilon)} \mathbb{E}_{(X,Y) \sim \eta(X|Y)\nu(Y)}l(f_{(\alpha, \beta)}(X),Y)   \nonumber \\
    &=& \mathbb{E}_{(X,Y) \sim \nu_{(1-\epsilon)\mu}(X|Y)\nu(Y)}l(f_{(\mu, 0)}(X),Y).
\end{eqnarray}
Here, $A(\nu_{\mu},\epsilon)= \{\eta(X|Y): \eta(\cdot|Y) $ is Gaussian,
$ W(\eta(X|Y),\nu_{\mu}(X|Y)) \leq \epsilon, \ \forall Y\in \{-1,1\}\}$.
\end{proposition}

This proposition, whose proof is discussed in Section~\ref{subsec:proof of prop DRO-Geodesic}.

\subsection{Data augmentation: worst-case interpolation
}
% {\color{blue}{AG: connect to te DRO result here}}
% {\color{red} Say the adversarial barycenter satisfies the dro setting}
In this section, we proceed to device computation strategies for the worst-case barycenter approach. Akin to robust training methods that seek adversarial examples approximately, such as PGD~\citep{madry2018towards} and FGSM~\citep{goodfellow2014explaining}, our paradigm learns an adversarial distribution~\citep{dong2020adversarial} and then samples from these distributions.

Here we make an important choice by focusing on the interpolation between $K=2$ distributions rather than a generalized $K$-margin barycenter.
%, like in ~\citep{fan2022generating_geodesic}. 
To date, efficient computational solutions for for multimarginal OT (MOT) are still nascent~\citep{lin2022complexity_multimargin_jmlr}. Solving an MOT with linear programming would scale to $O(n^K)$, and would be more costly when we want to solve for (free-support) barycenters to facilitate robustness in deep learning pipelines. Focusing on $K=2$ not only (1) allows us to focus on the decision boundary between the different categories of data, and even to deepen our understanding with theoretical analysis for robustness. (2) Achieve computation efficiency by leveraging established OT algorithms when iterating between pairs of distributions, as illustrated in Fig.(\ref{fig:fig1_illstration}).

Thus, our data augmentation strategy is to sample data from the worst-case adversarial Wasserstein barycenter $\{ \hat{x}_i, \hat{y}_i\} \sim \mu^{\{P_0,P_1\}}_{adv}$, which can be obtained as an inner maximization problem,
\begin{equation}\small
\begin{aligned}
    \mu^{\{P_0,P_1\}}_{adv} & :=  \max_{t \in [0, 1]} \mathbb{E}_{x, y \sim  \mu^{\{P_0,P_1\}}_{t} } \left[ l(f_\theta(x), y) \right]. %\\
    \label{eq:inner_max_da_adv_bc}
\end{aligned}    
\end{equation}
Despite a list of recent advances~\citep{villani2009optimal,korotin2022wasserstein_barycenter,fan2020scalable_barycenter}, the usage of Wasserstein barycenter may still be limited by scalability, and those solvers are more limited since we are optimizing the location of barycenters. However, thanks to the celebrated properties~\citep{brenier1991polar,mccann1997convexity}, which connects the Monge pushforward map with the coupling, we can avoid the intractable optimization over $\mu_t$ and turn to the training distribution with a transport map.

% Moreover, in practical problems such as image classification, we do not have access to the underlying data distributions $\{P_i\}_{i=1}^k$ but only the empirical distribution as the observed data samples $\{(x_i,y_i)\} \sim \sum w_i \tilde{P}_i$. 

%{\color{red} Tell the story: Simplify the problem from $\mathbb{E}_{p_t}$ to $\mathbb{E}_{p_0}$ in order to solve the problem}

\paragraph{Interpolation with OT map}
When $\nu_0,\nu_1 \in \mathcal{M}(\mathcal{X})$, $\nu_1$ is absolutely continuous with respect to $\nu_0$. Let $\mu(t) = \{\mu_t\}_{0\leq t \leq 1}$ be the geodesic of displacement interpolation, then we have a unique and deterministic interpolation $\mu_t = (T_t)\# P^0$, where $T_t(x):= (1-t)x + tT(x)$ and convex function $T(\cdot)$ is the optimal transport map~\citep{villani2009optimal}. 
Thus, the following proposition yields an objective for finding the worst-case barycenter:
% Thus, the transport is 
% \begin{equation}\small
% \begin{aligned}
% T_t(x) = (1 - t) x + t T(x),
% \label{eq:interpoalte_func}
% \end{aligned}
% \end{equation}
% where convex function $T(\cdot)$ is the optimal transport map~\citep{villani2009optimal}. We can have the following map:
\begin{proposition}
Consider a factored map $T^{xy}:=(T^x, T^y): \mathbb{R}^d \times \mathbb{R}^y \mapsto \mathbb{R}^d \times \mathbb{R}^y$ defined as  $T(\cdot,*):= (T^x(\cdot),T^y(*))$.
%, where $ T_t^x , T_t^y : \mathbb{R}^d \mapsto \mathbb{R}^d $. 
Consequently, rewriting Eq,~\eqref{eq:inner_max_da_adv_bc} with this above map yields
\begin{equation}\small
\begin{aligned}
    \tilde{\mu}^{\{P_0,P_1\}}_{adv} := & \max_{t \in [0, 1]} \mathbb{E}_{x, y \sim  P^0 } \left[ l(f_\theta(T^x_t(x)), T^y_t(y)) \right] \\
    \text{where, } & T^x_t(x) =(1 -t) x+ t T^x (x), \\
    & T^y_t(y) =(1 - t) y+ t T^y (y)
    \label{eq:adv_da_geo_interpolate}
\end{aligned}    
\end{equation}
such that the map 
%\TBD{$T^{xy}:= (T^x, T^y)$, rewrite it with a proposition} 
$T^{xy}$ satisfies the Monge problem $\min_{ T^{xy}\# P^0 = P^1} \mathbb{E}_{x,y \sim P^0} [ \mathcal{D}( \{x, y\}, T^{xy}(\{x, y\}) )^2]$. Here $\mathcal{D}(\{x_1,y_1\}, \{x_2, y_2\})=d(x_1 - y_1) + {L}(y_1,y_2)$ is a measure that combines the distance between samples the discrepancy between $y_1$ and $y_2$~\cite{courty2017joint_jdot}. We note here this joint cost is separable and the analysis of the generic joint cost function is left for future studies~\citep{fan2021scalable_map,korotin2022neuralot}.
\end{proposition} 

The benefits of this formulation: 
(1) Rather than the barycenter, the problem uses the OT map estimation where a lot of previous works have concerned efficiency~\citep{perrot2016mapping}, scalability~\citep{seguy2017large} and convergence~\citep{manole2021plugin_map_rate}. 
(2) The transport map can be estimated solely on data and stored regardless of the supervised learning task. Moreover, the map can project \textit{out-of-sample} points, which improves the generalization. 
(3) For the case where real data lies on a complex manifold, it is necessary to utilize an embedding space. Specifically, we perform the computation of Wasserstein geodesic and interpolation in an embedding space that is usually more regularized~\citep{kingma2013_vae}.
(4) Such OT results on labeled data distributions have been shown effective~\citep{alvarez2020geometric_otdd,alvarez2021dataset_otdd_flow,fan2022generating_geodesic} in transfer learning tasks while we first investigate its distributional robustness. Also, interestingly, when ignoring the expectation and defining the map as $T^x(x^0):=x^1$, 
$T^y(y^0):=y^1$ , such interpolation coincides with the popular mixup~\citep{zhang2018mixup} which is simple but effective even on complicated datasets.

\subsection{Implicit regularization: geodesic smoothness}
Data augmentation is, in general, viewed as implicit regularization~\citep{rebuffi2021data_aug_rob_deepmind}. However, our augmentation policy also motivates an explicit regularization term to enhance the training of deep neural networks. 

For a given predictive function $f^*_{\theta}: \mathbb{R}^d \mapsto \mathbb{R}^y$ and a loss criterion $l(\cdot, \cdot): \mathcal{Y} \times \mathcal{Y} \mapsto \mathbb{R}$, now we consider the following \textit{performance geodesic},  
\begin{equation}\small
\text{R}^{\text{g}}_{f^*_\theta,l} (t) :=  \mathbb{E}_{x, y \sim \mu(t)} \left[l ( f(x), y)
\right],
\end{equation}
which quantifies the loss associated with the prediction task, where $\mu(t)$ is the displacement interpolation as stated after Eq. ~\eqref{eq:inner_max_da_adv_bc}. 
The geodesic loss $\text{R}^{\text{geo}}_{f^*_\theta}$ provides us a new lens through which we can measure, interpret, and improve a predictive model's robustness from a geometric perspective.

% {\TBD{explain more about dynamic}}
Since the geodesic is a smooth and continuous path connecting $P^0,P^1$, The factorized geodesic interpolation $T^{xy}$ Eq. ~\eqref{eq:inner_max_da_adv_bc} allows us to formulate a new metric $\text{Reg}^{T^{xy}}_{l}$ that measures the change of a classifier $f^*_\theta$, under criteria $l(\cdot, \cdot)$ when gradually transporting from $P^0$ to $P^1$. %{\color{red} Modify}
%\TBD{Need a Lemma? }
%$\text{R}^{\text{geo}}_{f^*_\theta} (t)$ as follows 
{ \scriptsize
%\begin{equation}\small 
\begin{align}\small
    & \int_{0}^1  \left|  \frac{d}{dt} \int_{\mathcal{X} \times \mathcal{Y}} l ( f_\theta(T^x_t(x), T^y_t(y))
    d P^0(x,y) \right|  dt 
    =  \int_{0}^1 \left| \frac{d}{dt} \text{R}^{T^{xy}}_{f_\theta,l}(t) \right| dt 
    %:= \text{Reg}^{T^{xy}}_{l}(f_\theta)
    %& \text{where }  T_t(x) = (1-t)x + t T^* (x) 
\end{align}}
where $\text{R}^{T}_{f_\theta,l}(t) = \mathbb{E}_{x, y \sim P^0} \left[l( f^*_\theta(T^x_t(x)), T^y_t(y))\right]$ is the expected loss of $f^*_\theta$ at the location $t$ on the geodesic interpolation.
% \TBD{Illustration?} For example, ...
% It is usually those samples on the decision boundary, or the decision boundary itself that matters.
Thus, to robustify a classifier $f_\theta$, we propose to use the following regularization, which promotes the \textit{smoothness derivative along the geodesic}. 
{\small
\begin{align}
    \text{Reg}^{T}_{l}(f_\theta)
    := \int_{0}^1 \left|  \frac{d}{dt} \text{R}^{T}_{f_\theta,l}(t) \right| dt, \text{ (geodesic regularizer) }
\end{align}} 

\begin{proposition}[Geodesic performance as data-adaptive regularization]
Consider the following minimization 
    \begin{equation}
    \min_{f \in \mathcal{H}} \mathbb{E}_{x,y \sim P}[l(f(x), y)] + \frac{\lambda_1}{2} \text{Reg}^{\hat{T}}_l(f)^2 + \frac{\lambda_2}{2} \| f\|^2_{\mathcal{H}}.
    \nonumber
\end{equation}
where $\mathcal{H}$ is space of function.
When the data distribution is $P:=1/2P_0 + 1/2P_1$ and $P_0=\mathcal{N}(-\mu, I), P_1=\mathcal{N}(\mu,T)$.
The objective has the following form:
\begin{equation}
    \min_\theta \mathcal{L}(\theta) := - \langle \theta, \mu \rangle + \frac{\lambda_1}{2} | \langle \theta, \mu \rangle |^2 + \frac{\lambda_2}{2} \| \theta\|^2_2 \nonumber
\end{equation}
The optimal solution $f$ corresponding to data assumption has the closed form $\theta^* = (\lambda_1 \mu \otimes \mu + \lambda_2 I_m)^{-1} \mu \nonumber,$
% \begin{equation}
%     \theta^* = (\lambda_1 \mu \otimes \mu + \lambda_2 I_m)^{-1} \mu \nonumber,
% \end{equation}
where $\otimes$ refers to the outer product.
\end{proposition}
The illustrative result above indicates our geodesic regularizer is a \textit{data-depend regularization adjusted by the underlying data distributions}. This agrees with the recent theoretical understanding of mixup and robustness~\citep{zhang2021mix_rob_gen}.
%related to regularizing $\nabla^2 f_{\theta}$
%Connection with \citep{zhang2021mix_rob_gen}, and the regularization effect of Dropout training. 
In addition, existing robustness criteria (empirical and certified robustness)
%empirical robustness and certifiable \TBD{Need citation and intro in previous parts} robustness are two primary objectives for measuring robustness of neural networks. However, both metrics 
focus on the local area of individual samples within a perturbation ball $ \mathcal{S} = \{\delta: \| \delta \|_\infty \leq \epsilon, \epsilon > 0 \}$ without considering the robustness behavior of $f_\theta$ while transitioning from $P^0_{x,y}$ to $P^1_{x,y}$. The interpolation provides a new perspective for robustness, as shown in Fig.(\ref{fig:geodesic}).
% Both data augmentation and regularization have their own advantages and disadvantages. \TBD{Extend..}
\subsection{Computation and implementation}
\paragraph{OT map estimation} 
Our formulation allows us to leverage off-the-shelf OT map estimation approaches. 
Despite recent scalable and efficient solutions that solve the Monge problem ~\citep{seguy2017large}, withGAN~\citep{arjovsky2017wasserstein}, ICNN~\citep{makkuva2020optimal_icnn}, flow~\citep{huang2020convex_ot_flow,makkuva2020optimal_icnn,fan2021scalable_map}, and so on~\citep{korotin2022neuralot,bunne2022supervised_monge_map}.
As the first step, in this work, we refer to barycentric projection~\citep{ambrosio2005gradient_bc_proj} thanks to its flexibility~\citep{perrot2016mapping} and recent convergence understandings~\citep{manole2021plugin_map_rate,pooladian2021entropic_map_rate}. Specifically, we are not given access to the probability distributions $P^0$ and $P^1$, but only samples $\{(x^0_i, y^0_i) \}_{i=1}^n \sim P^0_{xy}$ and $\{(x^1_i, y^1_i) \}_{i=1}^n  \sim P^1_{xy}$. When an optimal map $T_0 \in \{T: \mathcal{X} \times \mathcal{Y} \mapsto \mathcal{X} \times \mathcal{Y}| T\#P^0 = P^1 \}$ exists, we want an estimator $\hat{T}_n$ with empirical distributions $\hat{P}^0_{xy,n}$ and $\hat{P}^1_{xy,n}$. This problem is usually regarded as the two-sample estimation problem ~\citep{manole2021plugin_map_rate}. 
Among them, the celebrated Sinkhorn algorithm~\citep{cuturi2013sinkhorn} solves the entropic Kantorovich objective efficiently $\hat{\pi}{\epsilon,n}:=\arg\min_{\pi \in \Pi (\mathbb{R}^{n \times n})} \langle \pi, M \rangle + \epsilon H(\pi)$, where $H(\pi):=-\sum_{ij} \pi_{ij} \log \pi_{ij}$ is the negative entropy, provides an efficient 
$O(n^2)$ solution. Furthermore, It was shown in Section 6.2 of \citep{pooladian2021entropic_map_rate} that an entropy-regularised map (given by $\hat{T}_{\epsilon,{n}}$) for empirical distributions can approximate the population transportation map efficiently.

% \begin{theorem}
% {\color{red} 
% \TBD{\citep{pooladian2021entropic_map_rate} section 6.2}
% } Let $\pi_\epsilon$ be the optimal entropic plan between empirical distributions $\hat{P^0}$ and $\hat{P^1}$, 
% \begin{equation}\small
% \begin{aligned}
%     \mathbb{E} \|\hat{T}_{\epsilon,{n}} - T_{0} \|^2_{L^2} \leq (1 + I_0(P^0,P^1)) \log(n) O(n^{-1/2d}),
% \end{aligned}    
% \end{equation}

% where $T_{\epsilon,n}(x)$ is the entropic-regularized map, and \TBD{...} 
% \end{theorem}
From the computation perspective, $T_{\epsilon,n}$ can be realized through the entropically optimal coupling~\citep{cuturi2013sinkhorn} $\pi^*_e$ as $\hat{T}_{\epsilon,n}(X_0) = \text{diag} (\pi^* \mathbf{1}_{n_1})^{-1} \pi^*_\epsilon X_1$. Since we assume the samples are drawn i.i.d, then   $\hat{T}_{\epsilon,n} (X_0) = n_0 \pi^* X_1$. At this point, we can have the geodesic interpolation computed via regularized OT as
\begin{equation}\small
\begin{aligned}
\hat{T}^{xy}_{t}: = 
\begin{cases}
\hat{T}^{x}_{t}(X_0) &= ( 1 - t )X_0 + t \hat{T}_{\epsilon,n} (X_0), % t n \pi^*_\epsilon X_1  
\\ 
% & = ( 1 - t )X_0 + t n \pi^*_\epsilon X_1 \\
\hat{T}^{y}_{t}(Y_0) &= (1 - t) Y_0 + t Y_1
\label{eq:map_computation}
\end{cases}
\end{aligned}
\end{equation}
%And the geodesic interpolation function becomes $T_t(x) = (1 - t) X_0 + t n_0 \pi^* X_1$. 
Since the labels $Y^i$ have a Dirac distribution within each $P_i$, we utilize the map $T^y(y^0):= y^1$. With such closed-form interpolation, we then generate samples by optimizing Eq.~\eqref{eq:adv_da_geo_interpolate} using numerical methods such as grid search. %{\color{red} talk about numerical}
%\TBD{actually, grid search..}.
Notice that, when setting $T^x(X_0) := X_1$ then the geodesic turns into the mixup however it can not deal with the case where we have different numbers of samples $n_0 \neq n_1$. 

% {\color{red} More contents here? regarding the regularization?}
\paragraph{Regularization} With the aforementioned parameterization,  we thus have the following expression for the smoothness regularizer  $\text{Reg}^{T}_{l}(f_\theta)$. 
% \begin{equation}\small
%     \begin{aligned}
%     \text{Reg}^{\hat{T}}_{l}(f_\theta) = \int_0^1 \left| \frac{d}{dt} \text{R}^{T^{xy}}_{f_\theta,l}(t)  \right| dt {\text{, {} where {} }} \frac{d}{dt} \text{R}^{T^{xy}}_{f_\theta,l}(t) 
%      = 
%     \int
%     \frac{dL}{d \tilde{y}} \nabla f_\theta ((1-t)x + t \hat{T}_{\epsilon,n} (x) )(\hat{T}_{\epsilon,n} (x) - x) d P^0(x,y) \\ 
%     %&= \frac{1}{n} \frac{dL}{d \tilde{y}} \nabla f_\theta ((1-t)X_0 + t n \pi^*_\epsilon X_1)(n \pi^*_\epsilon X_1 - X_0) %\\
%     % &\int_p^1 \left| \frac{dL}{d \tilde{y}} \nabla f_\theta ((1-t)X_0 + t n \pi^*_\epsilon X_1)(n \pi^*_\epsilon X_1 - X_0) \right| dt, 
%     \end{aligned}
% \end{equation}
\begin{equation}\small
    \begin{aligned}
    &\text{Reg}^{\hat{T}}_{l}(f_\theta) = \int_0^1 \left| \frac{d}{dt} \text{R}^{T^{xy}}_{f_\theta,l}(t)  \right| dt \text{ {} where {} } \frac{d}{dt} \text{R}^{T^{xy}}_{f_\theta,l}(t) \\
     & = 
    \int
    \frac{dL}{d \tilde{y}} \nabla f_\theta ((1-t)x + t \hat{T}_{\epsilon,n} (x) )(\hat{T}_{\epsilon,n} (x) - x) d P^0(x,y) \\ 
    \end{aligned}
\end{equation}
where $\tilde{y} = f_\theta((1-t)x + t\hat{T}(x))$ is the predicted target, $L=l(\tilde{y}, \hat{T}^y_t(y))$ measures the loss with regard to the interpolated target. This regularizer can be easily computed with the Jacobian of $f_\theta$ on mixup samples. In fact, we regularize the inner product of the expected loss, the Jacobian on interpolating samples, and the difference of $\hat{T}_{\epsilon,n}(x_0) - x_0$. In fact, Jacobian regularizer has already been shown to improve adversarial robustness~\citep{hoffman2019robust_jacob_adv}

\paragraph{Manifold and feature space}
The proposed data augmentation and regularization paradigms using the OT formulation have nice properties when data lie in a Euclidean space. 
%\TBD{quadratic ground cost, explanation}. 
However, the real data may lie on a complex manifold. In such a scenario,
we can use an embedding network $\phi: \mathcal{X} \mapsto \mathcal{Z}$ that projects it to a low-dimensional Euclidean space $Z \in \mathbb{R}^z$ and a decoder network $\psi: \mathcal{Z} \mapsto \mathcal{X}$. Similarly, data augmentation may be carried out via
\begin{equation}\small
    \begin{aligned}
    \tilde{\mu}^{\{P_0,P_1\}}_{adv} = & \max_{t \in [0, 1]} \mathbb{E}_{x, y \sim  P^0 } \left[ l(f_\theta( \psi( T^x_t(\phi(x) ))), T^y_t(y)) \right].
    \end{aligned}
\end{equation}
Also, regularize the geodesic on the manifold 
\begin{equation}\small
    \begin{aligned}\label{eq:comput_reg}
    &\text{Reg}^{\hat{T}}_{l}(f_\theta) = \int_0^1 \left| \frac{d}{dt} \text{R}^{T^{xy}}_{f_\theta,l}(t)  \right| dt {\text{, {} where }} \frac{d}{dt} \text{R}^{T^{xy}}_{f_\theta,l}(t) \\
    % & = 
    % \int
    % \frac{dL}{d \tilde{y}} \nabla f_\theta ((1-t)x + t \hat{T}_{\epsilon,n} (x) )(\hat{T}_{\epsilon,n} (x) - x) d P^0(x,y) \\ 
    &= \frac{1}{n} \frac{dL}{d \tilde{y}} \nabla f_\theta ((1-t) \phi(X_0)  + t n \pi^*_\epsilon \phi(X_1))(n \pi^*_\epsilon \phi(X_1) - \phi(X_0)). 
    \end{aligned}
\end{equation}
In fact, such a method has shown effectiveness in Manifold Mixup~\citep{verma2019manifold} since a well-trained embedding helps create semantically meaningful samples than mixing pixels, even on a Euclidean distance that mimics Wasserstein distance~\citep{courty2017learning_W_embedding}.
Recent advances in representation learning~\citep{radford2021learning_CLIP} also facilitate this idea.
In this work, we also adopt Variational Autoencoder (VAE)~\citep{kingma2013_vae} and pre-trained ResNets to obtain the embedding space. We follow a standard adversarial training scheme where we iteratively (a) update the predictor $f_\theta$ with objective $\min_\theta \mathbb{E}_{x_i,y_i \sim \sum P^i + \sum \nu_{adv} [l(f_\theta(x), y)]} + \lambda \text{Reg}(f_\theta)$, where the geodesic regularization $\text{Reg}(f_\theta)$ is computed via Eq.(\ref{eq:comput_reg}). (b) Find and store the augmented data by maximizing the equation (6). A pseudocode Algo.(2) is attached in the Appendix.

\paragraph{Scalability and batch OT} 
To handle large-scale datasets, we follow the concept of minibatch optimal transport~\citep{fatras2021minibatch_OT,fatras2019learning,khai2022improving_batch_OT} where we sample a batch of $b_n$ samples from only two classes during the data augmentation procedure. Whereas minibatch OT could lead to non-optimal couplings, it contributes to better domain adaptation performance~\citep{fatras2021unbalanced} without being concerned by the limitations of algorithmic complexity.
Further, our experimental results have demonstrated that our data augmentation is still satisfactory.  

% \begin{wrapfigure}{L}{0.45\textwidth}
% \begin{minipage}{0.99\linewidth}\small
\begin{algorithm}[H] %\big
    \centering
    \caption{The data augmentation algorithm}\label{alg:algorithm}
\begin{algorithmic}[1] %[1] enables line numbers
\FOR{ a batch of mixture data}
\STATE We only sample from two classes. $\{ X_{s,i}, y_{s,i}\}_{i=1}^{n_s}$, and $\{ X_{t,j}, y_{t,j}\}_{j=1}^{n_t}$. 
\STATE Map to manifold $Z_s = \phi_{rep}(X_s)$, $Z_t = \phi_{rep}(X_t)$. 
\STATE \textbf{\# Obtain the worst-case Wasserstein barycenters} 
\STATE Empirical distributions $\hat{\nu}_s = \sum_{i=1}^{n_s} p_{s,i} \delta_{Z_s}$, 
$\hat{\nu}_t = \sum_{j=1}^{n_t} p_{t,j} \delta_{Z_y}$. 
% \STATE Estimate  $\pi^*_\epsilon$ via Sinkhorn~\citep{cuturi2013sinkhorn}.
\STATE Sample a list of $\{ t_i\} \sim U[0, 1]$ for Monte-Carlo %estimation
\FOR{ each $t_i$}
\STATE \textbf{\# Get transported samples via \eqref{eq:map_computation}.}
\STATE Get the pushforward measures on the geodesic $\mu_{t_i} = (T_{t_i})\# \hat{\nu}_s$, 
\STATE Map to data space $\hat{X} = \phi^{-1}(\hat{Z}_s)$, $\hat{Z}_s \sim \mu_{t_i}$.
\STATE Prediction on the geodesic $\Tilde{y} = f(\hat{T}^x_{t_i} (\hat{X}))$, Compute the loss $L=l(\tilde{y}, \hat{T}^y_t(y))$.
\ENDFOR
\STATE Store the worst-case data $\hat{X}_{t_i}$ of this batch.
\ENDFOR
\STATE \textbf{return} All the augmented data
\end{algorithmic}
\end{algorithm}
% \end{minipage}
% \hfill
% \end{wrapfigure}

%%%%%%%%%%%%%%%%%%%%%%%%%%%%%%%%%%%%%%%%%%%%%%%%%%%%%
%%%%%%%%% Provable improvement of geodesic data aug
%%%%%%%%%%%%%%%%%%%%%%%%%%%%%%%%%%%%%%%%%%%%%%%%%%%%%%

\section{Theoretical analysis}
% In this section, we focus on investigating the benefits and properties of our proposed geodesic data augmentation. \TBD{Put some outlines here?}
\label{sec:theory}
Here, we rigorously investigate how the interpolation along geodesics affects the decision boundary and robustness.

We use the concrete and natural Gaussian model~\citep{schmidt2018adversarially} since it is theoretically tractable and, at the same time, reflects underlying properties in high-dimensional learning problems.
In fact, such settings  have been widely studied to support theoretical understandings in complex ML problems such as adversarial learning~\citep{carmon2019unlabeled_imp_rob,dan2020sharp_rob_gaussian}, self-supervised learning~\citep{deng2021improving_ood_data_rob_gaussian_mix}, and neural network calibration~\citep{zhang2022and_mix_cali_gaussian}. More importantly, recent advances in deep generative models such as GAN~\citep{goodfellow2020generative}, VAE~\citep{kingma2013_vae}, and diffusion model~\citep{song2019generative_score} endorse the theoretical analysis of Gaussian generative models~\citep{zhang2021mix_rob_gen} on complicated real-world data, akin to our manifold geodesic setting. 
% To provide tractable analysis, 
% \TBD{What's that?}
% w tractability in theory,s

\paragraph{Problem set-up} As introduced in our preliminary (Sec. \ref{sec:prelim}) and problem formulation (Sec. \ref{sec:adv_interpolate}), we consider a $2-$ classification task where all the data $S = \{ (x_1, y_1),...,(x_n, y_n)\} \sim \mathcal{P}^{all}_{x,y}$ are sampled from a joint distribution
$\mathcal{P}^{all}_{x,y} = 1/2 P^0_{xy} + 1/2 P^1_{xy}$. Formally, 
\begin{definition}[Conditional Gaussian model]

 For $\mu \in \mathbb{R}^d$ and $\sigma \geq 0$, the model is defined as the distribution over $(x,y) \in \mathcal{X} \times \mathcal{Y}$, where $\mathcal{X}\subseteq \mathbb{R}^d, \mathcal{Y} :=\{ -1, 1\}$,
 \begin{equation}\small
     \begin{aligned}
     (x,y) \sim P^{all}_{xy} :&= p(y=-1) P^0_{xy} +  p(y=1) P^1_{xy}  \nonumber
     %\\
     %&= p(x|y=0)p(y=0) + p(x|y = 1)p(y=1) \nonumber 
     \end{aligned}
 \end{equation}
 where $P^0_{xy} = N(- \mu, \sigma^2 I)$, $P^1_{xy} = N(\mu, \sigma^2 I)$, and $p(y=1)=p(y=-1)=\frac{1}{2}$.
\end{definition}
The goal here is is to learn a  classifier $f_\theta: \mathcal{X} \mapsto \mathcal{Y}$ parameterized by $\theta$ which minimizes population classification error
\begin{equation}\small
    \begin{aligned}
    R_{\mu,\sigma}(f):= \mathbb{E}_{x,y \sim P^{all}_{xy}}[ \mathbb{I}(f_\theta(x) \neq y)].
    \label{eq:class_err}
    \end{aligned}
\end{equation}
%are interested in mapping input $x \in \mathcal{X} \in \mathbb{R}^d$ to label $y \in \mathcal{Y}$. 
And such classifier is estimated from  $n_0$ observed training data samples $\{x_i, y_i \}_{i=1}^{n_0} \sim P^{all}_{x,y}$. 
%The goal here is to learn a deterministic classifier $f_\theta: \mathcal{X} \mapsto \mathcal{Y}$ parameterized by $\theta$. 
\subsection{Geodesic between Gaussian distributions} 
% {\color{red} The original distribution $P^r_{xy}$, the augmented distribution $P^A_{xy}$ }
%\paragraph{Geodesic data augmentation for Gaussian model} 
The assumption of Gaussian distribution not only provides a tractable guarantee for classification but also allows us to employ desirable conclusions for optimal transport between Gaussian distributions, thanks to established studies~\citep{dowson1982frechet_Gaussian_W_dist,givens1984class_Gaussian_W_dist,knott1984optimal_GD_map}.  
More specifically, although the Wasserstein distance and the transport map between regular measures rarely admit closed-form expressions, the $W_2$-Wasserstein distance between Gaussian measures can be obtained explicitly ~\citep{dowson1982frechet_Gaussian_W_dist}.  Moreover, the Optimal transport map~\citep{knott1984optimal_GD_map} between two Gaussian distributions as well as the constant speed geodesics between them, termed McCann's interpolation have explicit forms~\citep{mccann1997convexity}. Please see the Appendix for more detail on this. 
% {\color{blue}{AG: Don't need a theorem for this. Try to minimize as far as possible stating external results as theorems/propositions in the paper. Cite the relevant paper and provide a brief description instead. Stating multiple theorems from external works may confuse the theoretical reader as to our specific contribution and also diminish our contributions. Theorems should mostly be self-contained in the paper and reserved for contributions of the specific paper unless absolutely needed}}

Given explicit forms of the augmented distribution, to proceed with the theoretical analysis, we construct the following data augmentation scheme: We always consider a symmetric pair of augmented Gaussian distributions $G^g_{t}$ and $G^g_{1-t}$, $t\in [0, \frac{1}{2})$. Further, 
since the marginal distributions of the Gaussian model are $P^1 = N(- \mu, \sigma^2 I)$ and $P_2 = N(+ \mu, \sigma^2 I)$, we can denote the augmented data distribution as a mixture of Gaussian distributions 
as $P^{A(t)}_{xy}:= \frac{1}{2} G^g_{t} + \frac{1}{2} G^g_{1-t}$ with augmented samples $\tilde{x}_i, \tilde{y}_i : x|y \sim N(r y \mu, \sigma^2 I), y \sim U\{-1, 1\} $, where $ r = 1 - 2t$.
% $t$ as $P^{1, t}_{X, Y}: x \sim N(t y \mu, \sigma^2 I), y \sim U\{-1, 1\} $. 

% {\color{red} The definition of $t$ changes. Now $t$ means the geodesic, and $\lambda = 1 - 2t$ takes the part of $t$ in following analysis }

% \subsection{Supervised learning in Gaussian generative models}

% \begin{theorem}
% (Standard accuracy)
% \end{theorem}

% \begin{theorem}
% ((Sample complexity gap for robust accuracy)
% \end{theorem}

\subsection{Data augmentation improves robustness, provably}
Here, we demonstrate that the data augmentation obtained via Wasserstein geodesic perturbation provably improves the robust accuracy probability. 
In this work, we consider the $\ell_p$ norm ball with $p=\infty$. Specifically,
we apply a bounded worst-case perturbation before feeding the sample to the classifier. Then, we recap the definition of the standard classification and
robust classification error.
\begin{definition}[Standard, robust, and smoothed classification error probability~\citep{schmidt2018adversarially,carmon2019unlabeled_imp_rob}]
~\label{definition:classification_error_prob}
Let the original data distribution be $P_{X,Y}$. The standard error probability for classifier $f_\theta : \mathbb{R}^d \mapsto \mathcal{Y}$ is defined is $\mathbb{PE}_a(f_\theta) = P_{(x, y) \sim P_{X,Y}}( f_\theta (x) \neq y)$. The robust classification error probability is defined as $\mathbb{PE}^{p, \epsilon}_a(f_\theta) = P_{(x, y) \sim P_{X,Y}}(\exists u' \in \mathcal{B}^p_\epsilon(x),  f_\theta (u') \neq y)$, where the perturbations in a $\ell_p$ norm ball  $\mathcal{B}^p_\epsilon(x):= \{ x' \in \mathcal{X} | \|x' - x \| \leq \epsilon \}$ of radius $\epsilon$ around the input. The smoothed classification error for certifiable robustness is defined as $\mathbb{PE}^{\sigma_s}_a(f_\theta) = P_{(x, y) \sim P_{X,Y}, 
 \delta \sim N(0, \sigma_s^2 I )}( f_\theta (x + \delta) \neq y)$.
\end{definition}

% \begin{theorem}
% {\color{red} The standard and $\ell_2$-robust acc results?}
% \end{theorem}

\paragraph{Remark:} In this setting, standard, robust, and smoothed accuracies are aligned~\citep{carmon2019unlabeled_imp_rob}. In other words, a highly robust standard classifier will (1) also be robust to $\ell_\infty$ perturbation and (2) will be a robust base classifier for certifiable robustness, as randomized smoothing is a \textit{Weierstrauss transform} 
of the deterministic base classifier.

% \subsection{Benefits from the geodesic augmentation}
%Consider the linear classifiers $f_\theta = \text{sign}(\theta^\top x)$. 
%We develop theoretical results pertaining to the scenario where augmented data is labelled. 
%Consider the setting when augmented data is labelled by a user-defined external mechanism.
%This scenario is same as considering the available data as consisting of both the original and augmented samples. 
The following result provides a bound for the robust error in such a case compared to using the original data alone.

\begin{theorem}
\label{theorem:Data_aug_Gaussian}
Suppose the original data $\{ {X}_i, {Y}_i, \}_{i=1}^{{n_0}}$ are iid with distribution satisfying $X|Y=y \sim N(y\mu, \sigma^2 I)$, $Y\sim U\{-1,1\}$. Additionally our $n_1 = m_a n_0$ augmented data $\{ \Tilde{X}_i, \Tilde{Y}_i\}_{i=1}^{n_1}$ are iid and independent of the original data and satisfies $X|Y=y \sim N (y r\mu, \sigma^2 I)$, $Y\sim U\{-1,1\}$ where $r \in [0, 1]$, are the Wasserstein geodesic interpolation. Then for $\epsilon>0$, for all $n_1>N_1$ with $N_1$ (depending on $t$) satisfying $tN_1 +n_0 >\sqrt{N_1} \log(N_1)$ and $\sqrt{n_0} (r+n_0)/(\sqrt{N_1}\sigma)\leq \log{N_1}$, then we have
% $
% \mathbb{PE}^{p, \epsilon}_a(f_{\hat{\theta}_{n_0}}) \geq   \mathbb{PE}^{p, \epsilon}_a(f_{\hat{\theta}_{n_0,n_1}}),
% $
\begin{equation}\small
\mathbb{PE}^{p, \epsilon}_a(f_{\hat{\theta}_{n_0}}) \geq    \mathbb{PE}^{p, \epsilon}_a(f_{\hat{\theta}_{n_0,n_1}}),
\end{equation}
\vspace{-10pt}
\begin{equation}\small
w.p. \geq P_{X\sim N(0, I_d)} (\|X\| \leq \log(n_1)/\sigma) \times P_{X\sim N(0, I_d)}( A \cap B  ),    
\end{equation}
where 
$\scriptsize{ A= \{X: \sqrt{n_0} (r+n_0)/(\sqrt{n_1}\sigma) \leq \|X\| \leq \log{n_1} \}}, $
$\scriptsize{B=\{X:\mu^TX/\|X\| \leq \sqrt{\|\mu\|^2-\{n_0 (r+n_0)^2/({n_1}\sigma^2)}\},}$ and $\small{\hat{{\theta}}_{n_0,n_1}={(\sum_{i=1}^{n_0}Y_i X_i + \sum_{i=1}^{n_1}\tilde{Y}_i \tilde{X}_i )} / {n_0 + n_1},}$
$\small{\hat{{\theta}}_{n_0}={\left(\sum_{i=1}^{n_0}Y_i X_i \right)} / {n_0 }}$.
% \begin{equation}\scriptsize
% \begin{aligned}
% A= \{X: \sqrt{n_0} (t+n_0)/(\sqrt{n_1}\sigma) \leq \|X\| \leq \log{n_1} \}, 
% B=\{X:\mu^TX/\|X\| \leq \sqrt{\|\mu\|^2-\{n_0 (t+n_0)^2/({n_1}\sigma^2)}\},
% \end{aligned}    
% \end{equation}
% \vspace{-10pt}
% \begin{equation}\small
% \begin{aligned}
% \hat{{\theta}}_{n_0,n_1}=\dfrac{\left(\sum_{i=1}^{n_0}Y_i X_i + \sum_{i=1}^{n_1}\tilde{Y}_i \tilde{X}_i \right)}{n_0 + n_1},
% \hat{{\theta}}_{n_0}=\dfrac{\left(\sum_{i=1}^{n_0}Y_i X_i \right)}{n_0 }
% \end{aligned}    
% \end{equation} 
$f_{\theta}(x)$ defined as $sign(x^T\theta)$.
\end{theorem}
Note that in the above theorem $P_{X\sim N(0, I_d)} (\|X\| \leq \log(n_1)/\sigma) \to 1$ as $n_1 \to \infty$. Moreover, $P(A)$ and $P(B) \to 1$ as well with $n \to \infty$. The proof is shown in the Appendix.
\paragraph{Remark:} 
From the above theory, we can see that the robustness from data augmentation will increase when we have a larger $r$ and sample size. This explains the property of interpolation-based methods where the samples from vicinal distributions are desired. Actually, the effect of sample size can be observed in the following experiments (figure (\ref{fig:ablation})).

%%%%%%%%%%%%%%%%%%%%%%%%%%%%%%%%%%%%%%%%
%%%%%%%%%% Experiments
%%%%%%%%%%%%%%%%%%%%%%%%%%%%%%%%%%%%%%%%

\section{Experiments and Discussion}\label{sec:exp}
We evaluate our proposed method in terms of both empirical robustness and certified robustness~\citep{cohen2019certified_cert_rob} on the MNIST \citep{LeCun1998GradientbasedLA}, CIFAR-10 and CIFAR-100~\citep{Krizhevsky2009LearningML} dataset, and samples from ImageNet($64\times64$) dataset~\citep{deng2009imagenet,le2015tiny_imagenet}. 
Typically, we use data augmentation to double the size of the training set $m_a = 1$ at first and use the regularization with a fixed weight $\alpha_{reg}=5.0$ during the training, as implicit data augmentation. For the MNIST samples, we train a LeNet~\citep{LeCun1998GradientbasedLA} classifier with a learning rate $lr=0.01$ for $90$ epochs. For the CIFAR dataset, we use the ResNet110~\citep{He2016DeepRL} for the certifiable task on CIFAR10 and PreactResNet18 on CIFAR-100. The Sinkhorn entropic coefficient is chosen to be $\epsilon=0.01$. We use VAE with different latent dimensions as embedding, where details can be found in the appendix. The experiments are carried out on several NVIDIA RTX A6000 GPUs and two NVIDIA GeForce RTX 3090 GPUs.

\begin{figure}[tb]
\minipage{0.4\textwidth}
  \includegraphics[width=0.9\linewidth]{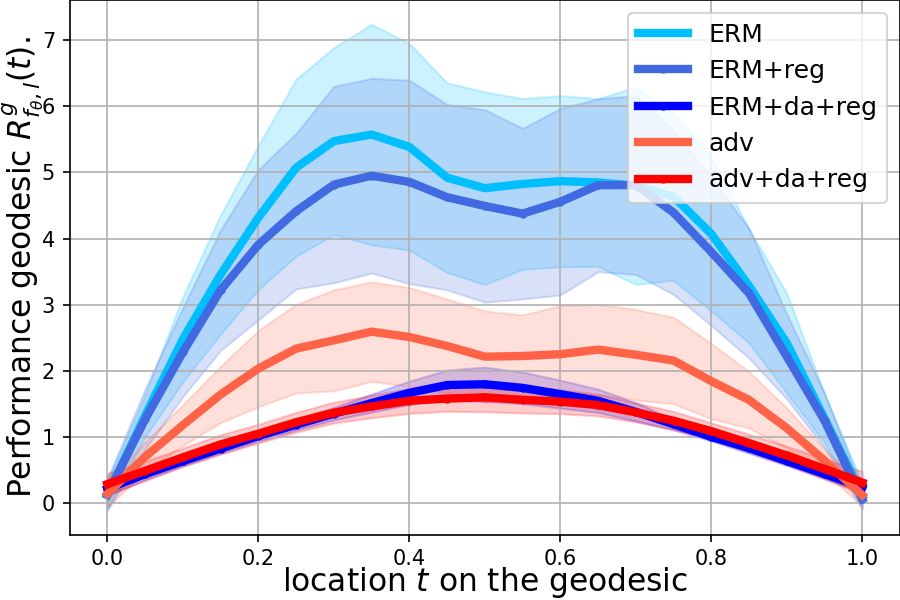}
  \caption{
The \textit{performance geodesic} $\text{R}^{\text{g}}_{f,l} (t)$ of classifiers obtained from different training strategies. }
\label{fig:geodesic}
\endminipage\hfill
\minipage{0.58\textwidth} 

\begin{adjustbox}{width=0.99\linewidth}
    \begin{tabular}{c|ccccc}
    \toprule
  & ERM  & PGD & ERM+\textbf{Reg} & ERM + \textbf{Ours} & PGD +\textbf{Ours}   \\ 
    \midrule
     
\text{Clean Acc.(\%)} &99.09 $\pm$ 0.02	 & 98.94 $\pm$ 0.03	& 99.27 $\pm$ 0.06 & \textbf{99.39 $\pm$ 0.03} & 99.34 $\pm$ 0.02 \\	
\text{Robust Acc.(\%)} & 31.47 	$\pm$ 0.12	& 81.23 $\pm$	0.17 & 35.66 $\pm$ 0.20 & 81.23 $\pm$ 0.12	& \textbf{82.74 $\pm$ 0.16}		\\

\bottomrule
\end{tabular}
    \end{adjustbox}
\vspace{-10pt}
\captionof{table}{Adversarial robustness (MNIST), ours:DA $+$ Reg
}
\label{table:pgd}

\begin{adjustbox}{width=0.99\linewidth}
    \begin{tabular}{c|ccccccc}
    \toprule
  & ERM  & Mixup & Manifold & CutMix & AugMix  & PuzzleMix & \textbf{Ours} \\ 
    \midrule
     
\text{Clean Acc.(\%)} & 58.28  & 56.32 	& 58.08  & 57.71  & 56.135 & 63.31 & \textbf{64.15} \\	
\text{FGSM Acc.(\%)} & 8.22		& 11.21	& 10.69 & 11.61	& 11.07 & 7.82 & \textbf{13.46} \\ \bottomrule
\end{tabular}
    \end{adjustbox}
    \vspace{-10pt}
\captionof{table}{$\ell_\infty$ Standard and adversarial robust accuracy on ImageNet $(64\times64)$ with baselines
}
\label{table:imagenet}

\begin{adjustbox}{width=0.99\linewidth}
\begin{tabular}{c|ccccccc}
    \toprule
  & ERM  & Mixup & Manifold & CutMix & AugMix  & PuzzleMix & \textbf{Ours} \\ 
    \midrule
     
\text{Clean Acc.(\%)} & 78.76	 & 81.90	& 81.98 & 82.31 & 79.82 & 83.05 & 81.36 \\	
\text{FGSM Acc.(\%)} & 34.72		& 43.21	& 39.22 & 30.81	& 44.82 & 36.82 & \textbf{54.87} \\ \bottomrule
\end{tabular}
\end{adjustbox}
\vspace{-10pt}
\captionof{table}{$\ell_\infty$ Standard and adversarial robust accuracy on CIFAR-100 dataset with baselines
}
\label{table:cifar100}

\endminipage\hfill
\end{figure}

\paragraph{Empirical robustness} 
% {\color{red}{AG: put remarks following the tables}}
We use the strongest PGD method to perform $\ell_\infty$ attack with $\epsilon = 2.0$ and $4$ steps. As shown in table (\ref{table:pgd}), either data augmentation and regularization can improve the $\ell_\infty$ robustness. Moreover, our method can further improve gradient-based adversarial training. In fig.(\ref{fig:geodesic}), we visualize the \textit{performance geodesic} of various training strategies where more robust models apparently have mode smoother geodesic. Here \textit{ERM} stands for normal training, \textit{$+$reg} is using geodesic regularization, \textit{$+$da} means using data augmentation, \textit{adv} denotes adversarial training with PGD. 
For empirical robustness on CIFAT-100 and ImageNet($64\times64$), we follow training protocol from ~\citep{kim2020_puzzlemix} and compare our method with ERM, vanilla mixup~\citep{zhang2018mixup}, Manifold mixup~\citep{verma2019manifold}, CutMix~\citep{yun2019cutmix}, AugMix~\citep{hendrycks2019_augmix}, and PuzzleMix~\citep{kim2020_puzzlemix}. To enable a fair comparison, we reproduce only the non-adversarial PuzzleMix methods.

% \begin{wraptable}{l}{90mm}
% % \begin{table}[H]\small
% % \centering
% % \vspace{-5pt}
%     \begin{adjustbox}{width=0.99\linewidth}
%     \input{tables/tiny_image_net}
%     \end{adjustbox}
%     \vspace{-5pt}
% \caption{$\ell_\infty$ Standard and adversarial robust accuracy on ImageNet $(64\times64)$ with baselines
% }
% \label{table:imagenet}

% \begin{adjustbox}{width=0.99\linewidth}
% \input{tables/cifar100.tex}
% \end{adjustbox}
% \vspace{-5pt}
% \caption{$\ell_\infty$ Standard and adversarial robust accuracy on CIFAR-100 dataset with baselines
% }
% \label{table:cifar100}
% % \end{table}
% % \vspace{-10pt}
% \end{wraptable}

For ImageNet($64\times64$), it contains 200 classes with $64\times64$ resolution\citep{chrabaszcz2017downsampled_tiny_imagenet}. As in table (\ref{table:imagenet}), our approach outperforms the best baseline by $16.8\%$ under ($\epsilon=4/255$) FGSM attack. For CIFAR-100, as in table (\ref{table:cifar100}), we exceed baseline by $22.4\%$ under FGSM, while having comparable performance in standard accuracy. It may be caused by our data augmentation populating samples only on the local decision boundaries.

% \paragraph{ImageNet$(64\times64)$}
% Samples from ImageNet, 200 classes with 500 training images and 50 test images 
% FGSM ($\epsilon=4/255$) \TBD{Finish the rest.}

% \begin{wraptable}{l}{90mm}
% % \begin{table}[H]\small
% % \centering
% \vspace{-5pt}
% \begin{adjustbox}{width=0.99\linewidth}
% \input{tables/cifar100.tex}
% \end{adjustbox}
% \caption{$\ell_\infty$ Standard and adversarial robust accuracy on CIFAR-100 dataset with baselines
% }
% \label{table:cifar100}
% % \end{table}
% \vspace{-10pt}
% \end{wraptable}

% \begin{wrapfigure}{l}{100mm}
\begin{figure}[t]
	\centering
\includegraphics[width=0.65\linewidth]{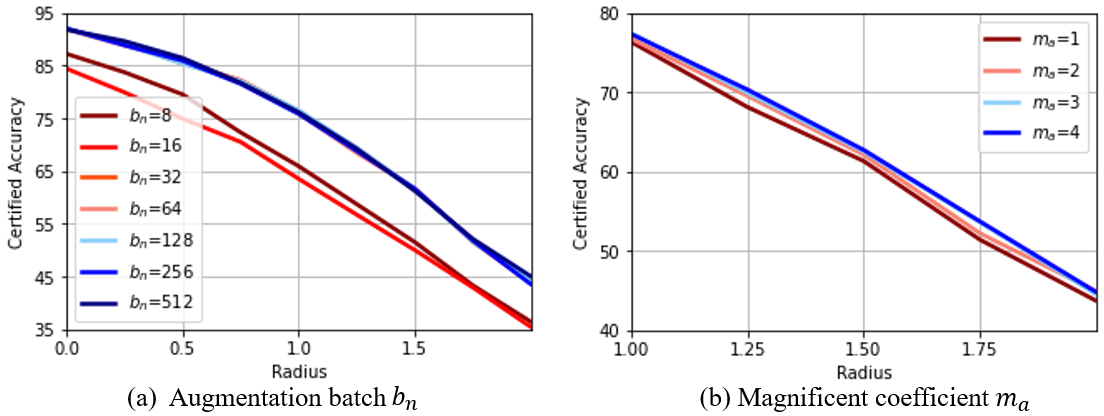}
	\caption{Comparison of approximate certified accuracy with different parameters, including (a) augmentation batch size $b_n$, and (b) magnificent coefficient $m_a$.}
	\label{fig:ablation}
\end{figure}
% \end{wrapfigure} 

\paragraph{$\ell_2$ Certifiable robustness}We compare the performance with Gaussian \citep{cohen2019certified_cert_rob}, Stability training \citep{li2018certified_renyi}, SmoothAdv \citep{salman2019provably}, MACER \citep{Zhai2020MACER}, Consistency \citep{jeong2020consistency}, and SmoothMix \citep{jeong2021smoothmix}. 
We follow the same evaluation protocol proposed by \citep{cohen2019certified_cert_rob} and used by previous works~\citep{salman2019provably,Zhai2020MACER,jeong2020consistency}, which is a Monte Carlo-based certification procedure, calculating the prediction and a "safe" lower bound of radius over the randomness of $n$ samples with probability at least $1-\alpha$, or abstaining the certification.
We consider three different models as varying the noise level $\sigma \in \{0.25, 0.5, 1.0 \}$. During inference, we apply randomized smoothing with the same $\sigma$ used in training. The parameters in the evaluation protocol \citep{cohen2019certified_cert_rob}  are set as: $n=100,000$, $n_0 = 100$, and $\alpha = 0.001$,  with previous work \citep{cohen2019certified_cert_rob,jeong2020consistency}.

% \subsection{$\ell_2$ Certifiable robustness}

\begin{table}[tp]\small
\centering
\caption{Certified accuracy on CIFAR-10 dataset. %\TBD{Directly show the best result, add more variant of ours}
}
\vspace{-0.03in}
    \begin{adjustbox}{width=0.9\linewidth}
    \begin{tabular}{cl|cccccccccc}
    \toprule
    $\sigma$ &  Models (CIFAR-10)  & 0.00 & 0.25 & 0.50 & 0.75 & 1.00 & 1.25 & 1.50 & 1.75 & 2.00 & 2.25 \\ 
    \midrule
    \multirow{7.5}{*}{0.25}& Gaussian   & {76.6} & {61.2} & {42.2} & {25.1} & 0.0 & 0.0 & 0.0 & 0.0  & 0.0 & 0.0 \\
    & Stability training   & 72.3 & 58.0 & 43.3 & 27.3 & 0.0 & 0.0 & 0.0 & 0.0  & 0.0 & 0.0 \\ 
    & SmoothAdv$^*$  & {73.4} & {65.6} & {57.0} & {47.1} & 0.0 & 0.0 & 0.0 & 0.0   & 0.0 & 0.0\\
    & MACER$^*$     & {79.5} & {69.0} & {55.8} & {40.6} & 0.0 & 0.0  & 0.0 & 0.0 & 0.0 & 0.0\\
    & Consistency   & 75.8 & {67.6} & {58.1} & {46.7} & 0.0 & 0.0 & 0.0   & 0.0 & 0.0 & 0.0\\
    & {SmoothMix }  & {77.1} & {67.9} & {57.9} & {46.7} & 0.0 & 0.0 & 0.0  & 0.0 & 0.0 & 0.0\\
    \cmidrule(l){2-2} \cmidrule(l){3-3} \cmidrule(l){4-12}
    & \textbf{Ours} &69.7	&\bf{63.4}	&\bf{56.0}	&\underline{\bf{47.3}}  & 0.0 & 0.0 & 0.0 & 0.0 & 0.0  & 0.0\\
    \midrule
    \multirow{7.5}{*}{0.50}& Gaussian   & {65.7} & {54.9} & {42.8} & {32.5} & {22.0} & {14.1} & {8.3} & {3.9}  & 0.0 & 0.0\\
    & Stability training   & 60.6 & 51.5 & 41.4 & 32.5 & 23.9 & 15.3 & 9.6 & 5.0 & 0.0 & 0.0\\ 
    & SmoothAdv$^*$   & 65.3 & 57.8 & 49.9 & 41.7 & 33.7 & 26.0 & 19.5 & 12.9 & 0.0 & 0.0\\  
    & MACER$^*$   & {64.2} & {57.5} & {49.9} & {42.3} & {34.8} & {27.6} & {20.2} & {12.6}  & 0.0 & 0.0\\
    & Consistency    & 64.3 & {57.5} & {50.6} & {43.2} & {36.2} & {29.5} & {22.8} & {16.1} & 0.0 & 0.0\\
    & {SmoothMix }  & 65.0 & {56.7} & {49.2} & {41.2} & {34.5} & {{29.6}} & {{23.5}} & {{18.1}}& 0.0 & 0.0 \\
    \cmidrule(l){2-2} \cmidrule(l){3-3} \cmidrule(l){4-12}
    & \textbf{Ours} &55.4	&50.7	&\bf{45.6} &	\bf{40.6}	&\bf{35.2}	&\underline{\bf{30.5}}	&\underline{\bf{25.7}}	&\underline{\bf{19.5}}	&0.0	&0.0	 \\
    \midrule
    \multirow{7.5}{*}{1.00}& Gaussian   & 47.2 & 39.2 & 34.0 & 27.8 & 21.6 & 17.4 & 14.0 & 11.8 & 10.0 & 7.6 \\
    & Stability training   & 43.5 & 38.9 & 32.8 & 27.0 & 23.1 & 19.1 & 15.4 & 11.3 & 7.8 & 5.7 \\
    & SmoothAdv$^*$& 50.8 & 44.9 & 39.0 & 33.6 & 28.1 & 23.7 & 19.4 & 15.4 & 12.0 & 8.7  \\ 
    & MACER$^*$    & {40.4} & {37.5} & {34.2} & {31.3} & {27.5} & {23.4} & {22.4} & {19.2} & {16.4} & {13.5}  \\
    & {Consistency}   & 46.3 & {41.8} & {37.9} & {34.2} & {30.1} & {26.1} & {22.3} & {19.7} & {16.4} & {13.8} \\
    & {SmoothMix}   & 47.1 & {42.5} & {37.5} & {32.9} & {28.2} & {24.9} & {21.3} & {18.3} & {15.5} & {12.6} \\
    \cmidrule(l){2-2} \cmidrule(l){3-3} \cmidrule(l){4-12}
    & \textbf{Ours} &40.9	&36.9	&\bf{34.5}	&\bf{31.5}	&\bf{28.1}	&\underline{\bf{24.4}} &	\underline{\bf{22.5}}	&\underline{\bf{20.4}}	&\underline{\bf{16.5}}	&\underline{\bf{14.0}}	%&12.3 
    \\
    \bottomrule
\end{tabular}
    \end{adjustbox}
\label{table:cifar10}
\end{table}

For the CIFAR-10 dataset, Table~\ref{table:cifar10} showed that our method generally exhibited better certified robustness compared to other baselines, i.e.,  SmoothAdv \citep{salman2019provably}, MACER \citep{Zhai2020MACER}, Consistency \citep{jeong2020consistency}, and SmmothMix \citep{jeong2021smoothmix}. The important characteristic of our method is the robustness under larger noise levels. Our method achieved the highest certified test accuracy among all the noise levels when the radii are large, i.e., radii 0.25-0.75 under noise level $\sigma=0.25$, radii 0.50-1.75 under noise level $\sigma=0.50$, and radii 0.50-2.75 under noise level $\sigma=1.00$. We obtained the new state-of-the-art certified robustness performance under large radii.  
%We also combined our method with SmoothAdv and SmoothMix to evaluate whether our data augmentation method can provide additive improvements. 
As shown in Table~\ref{table:cifar10}, we find that by combing our data augmentation mechanism, the performance of previous SOTA methods can be even better, which demonstrates our method can be easily used as an add-on mechanism for other algorithms to improve robustness.

\paragraph{Ablation study} Here, we conducted detailed ablation studies to investigate the effectiveness of each component. We performed experiments on MNIST with $\sigma=1.0$, and the results are shown in Figure~\ref{fig:ablation}. (More results are shown in the Appendix.) We investigated the influence by the augmentation batch size $b_n$ (Figure~\ref{fig:ablation}(a)) following the batch optimal transport setting. Also, we studied data augmentation sample size  $m_a$ (Figure~\ref{fig:ablation}(b)). 
% {\color{red} Change}
We found that a larger augmentation batch size $b_n$ leads to better performance, which is expected since it better approximates the joint measure. 
In addition, more augmented data samples $m_a$ benefit robustness, which agrees with our theoretical results.

\section{Conclusion and Future Work} {\color{red}}
In this report, we proposed to characterize the robustness of a model through its performance on the Wasserstein geodesic connecting different training data distributions. The worst-case distributions on the geodesic allow us to create augmented data for training a more robust model. Further, we can regularize the smoothness of a classifier to promote robustness.
We provided theoretical guarantees and carried out extensive experimental studies on multiple datasets, including CIFAR100 and ImageNet. 
Our results showed new SOTA performance compared with existing methods and can be easily combined with other learning schemes to boost the robustness. 

As a first step, this work provides a new perspective to characterize the model's robustness. There could be several future works, including considering multi-marginal adversarial Wasserstein barycenter on a simplex, more efficient optimization on the geodesic, and more thorough theoretical studies beyond Gaussian models.

\bibliography{reference}
\bibliographystyle{tmlr}

%%%%%%%%%%%%%%%%%%%%%%%%%%%%%%%%%%%%%%%%%%%%%%%%%%%%%%%%%%%%%%%%%%%%%%%%%%%%%%%
%%%%%%%%%%%%%%%%%%%%%%%%%%%%%%%%%%%%%%%%%%%%%%%%%%%%%%%%%%%%%%%%%%%%%%%%%%%%%%%
% APPENDIX
%%%%%%%%%%%%%%%%%%%%%%%%%%%%%%%%%%%%%%%%%%%%%%%%%%%%%%%%%%%%%%%%%%%%%%%%%%%%%%%
%%%%%%%%%%%%%%%%%%%%%%%%%%%%%%%%%%%%%%%%%%%%%%%%%%%%%%%%%%%%%%%%%%%%%%%%%%%%%%%
%\clearpage
\appendix

\onecolumn

\addcontentsline{toc}{section}{Appendix} % Add the appendix text to the document TOC
\part{Appendix} % Start the appendix part
\parttoc % Insert the appendix TOC

% \section{Additional introduction}

\section{Theoretical results}

\subsection{The Wasserstein distance between Gaussian distribution }

\begin{theorem}[The Wasserstein distance between Gaussian distribution 
%{\color{red} Do we need two-sample asymptotics? No.. not that related}
]

\label{theorem:Wasserstein-Gaussian}
Given two Gaussian measures, $P^0 = N(\mu_0, \Sigma_0)$ and $P^1 = N(\mu_1, \Sigma_1)$. The $L^2$-Wasserstein distance between them is given by
\begin{equation}\small
\begin{aligned}
W^2_2(P^0, P^1) = & \|\mu_0 - \mu_1 \|^2 + tr(\Sigma_0) + tr(\Sigma_1) \\
&- 2 tr (\sqrt{\Sigma_0^{1/2} \Sigma_1 \Sigma_0^{1/2}})
\end{aligned} 
\end{equation}
% Consider i.i.d samples $\{x^0_1,..,x^0_n\}$ and $\{x^1_1,..,x^0_m\}$
\label{theorem:w_gaussian}
\end{theorem}
The proof largely depends on a summary of \cite{givens1984class_Gaussian_W_dist}, as well as. 
We recall the theorem. Given two Gaussian measures, $P^0 = N(\mu_0, \Sigma_0)$ and $P^1 = N(\mu_1, \Sigma_1)$. The $L^2$-Wasserstein distance between them is given by
\begin{equation}
\begin{aligned}
W^2_2(P^0, P^1) = & \|\mu_0 - \mu_1 \|^2 + tr(\Sigma_0) + tr(\Sigma_1) 
- 2 tr (\sqrt{\Sigma_0^{1/2} \Sigma_1 \Sigma_0^{1/2}})
\end{aligned}
\end{equation}
Given two distributions $P^0 = N(\mu_0, \Sigma_0)$ and $P^1 = N(\mu_1, \Sigma_1)$, one can first reduce to the centered case $\mu_0 = \mu_1 = 0$. Next, let $X \sim P^0$ and $Y \sim P^1$ and if $(X, Y)$ is a random vector of $\mathbb{R}^n \times \mathbb{R}^n$ with covariance matrix
\begin{equation*}
    \Gamma = 
    \begin{bmatrix}
    \Sigma_0 & C \\
    C^\top & \Sigma_1
    \end{bmatrix}
\end{equation*}
then the quantity
\begin{equation*}
    \mathbb{E}[ \| X - Y \|^2_2] = Tr (\Sigma_0 + \Sigma_1 - 2 C)
\end{equation*}
depends only on $\Gamma$. Also, when $P^0 = N(0, \Sigma_0)$ and $P^1 = N(0, \Sigma_1)$, one can restrict the infimum which defines $W_2$ to run over $W_2$ to run over Gaussian laws $N(0, \Gamma)$ on $\mathbb{R}^n \times \mathbb{R}^n$ with covariance matrix $\Gamma$ structured above. Here the constrain on $C$ is the Schur complement constraint $\Sigma_0 - C \Sigma^{-1}_1 C^\top \succeq 0.$ Thus, the minimization of the function $C \mapsto -2 Tr(C)$ leads to the result.

\subsection{Transport map between Gaussian measures }

\begin{proposition}[Optimal transport map between Gaussian measures, adopted from ~\cite{knott1984optimal_GD_map}, Example 1.7~\cite{mccann1997convexity}]
% Having two Gaussian distributions $P^1 = N(\mu_1, \Sigma_1)$ and $P_2 = N(\mu_2, \Sigma_2 )$ in $\mathcal{P}(\mathbb{R}^d)$ respectively, 
Given the two Gaussian distributions defined in theorem (\ref{theorem:w_gaussian}), 
define a symmetric positive matrix $T$ and a corresponding linear map $\mathcal{T} : \mathcal{X} \mapsto \mathcal{X}$ 
\begin{equation}\small 
\begin{aligned}
\mathcal{T}(x) = \mu_1 + T (x - \mu_0), { {} where }\\
T = \Sigma_1^{1/2}(\Sigma_1^{1/2}\Sigma_0\Sigma_1^{1/2})^{-1/2}\Sigma_1^{1/2}
\end{aligned}
\end{equation}
Then, $\mathcal{T}$ is the optimal map such that $P^1 = T\#P^0$.
\end{proposition}
Hence we obtain a closed-form map and that leads to an explicit form of the geodesic interpolation among two Gaussian distributions, as:
\begin{equation}\small
\begin{aligned}
{T}^{G,xy}_{t}: =  \nonumber
\begin{cases}
{T}^{G,x}_{t}(x^0) &= ( 1 - t ) x^0 + \mathcal{T} (x^0),   
\\ 
{T}^{G,y}_{t}(y^0) &= \begin{cases}
y^0, { {} } 0 < t < 0.5, \\
-y^0, { {} } 0.5 < t < 1.
\end{cases}
\end{cases}
\end{aligned}
\end{equation}
Here the interpolation of $y^0$ means we will always label a sample as $-1$ or $1$, which allows us to proceed to the analysis under the classification error.
Given the two Gaussian distributions defined in theorem (\ref{theorem:w_gaussian}), 
define a symmetric positive matrix $T$ and a corresponding linear map $\mathcal{T} : \mathcal{X} \mapsto \mathcal{X}$ 
\begin{equation}\small 
\begin{aligned}
\mathcal{T}(x) = \mu_1 + T (x - \mu_0), { {} where }\\
T = \Sigma_1^{1/2}(\Sigma_1^{1/2}\Sigma_0\Sigma_1^{1/2})^{-1/2}\Sigma_1^{1/2} \nonumber
\end{aligned}
\end{equation}
Then, $\mathcal{T}$ is the optimal map such that $P^1 = T\#P^0$.

Check the optimal transport map from $N(\mu_0, \Sigma_0)$ to $N(\mu_1, \Sigma_1)$, again assume $\mu_0= \mu_1 = 0$ for simplicity, one may define the random column vectors $X \sim N(\mu_0, \Sigma_1)$ and $Y = T X$ and write
\begin{equation}
    \begin{aligned} \nonumber
    \mathbb{E} (YY^\top) &=  T \mathbb{E} (XX^\top)T^\top \\
    &= \Sigma_0^{-1/2}(\Sigma^{1/2}_1 \Sigma_2 \Sigma^{1/2}_0)^{1/2} (\Sigma^{1/2}_1 \Sigma_2 \Sigma^{1/2}_0)^{1/2} \Sigma_0^{-1/2} \\
    &= \Sigma_1
    \end{aligned}
\end{equation}
To show the map is optimal, one may use
\begin{equation}
    \begin{aligned} \nonumber
    \mathbb{E}(\| X - Y\|^2_2) &= \mathbb{E} (\|X \|^2_2) + \mathbb{E} (\| Y \|^2_2) - 2 \mathbb{E} (<X, Y>) \\
    &= Tr(\Sigma_0) + Tr(\Sigma_1) - 2 \mathbb{E}(<X, TX>) \\
    &= Tr(\Sigma_0) + Tr(\Sigma_1) - 2 Tr(\Sigma_0 T)
    \end{aligned}
\end{equation}
and one can observe that by the cyclic property of the trace
\begin{equation}
    Tr(\Sigma_0 T) = Tr((\Sigma^{1/2}_1 \Sigma_2 \Sigma^{1/2}_0)^{1/2}) \nonumber
\end{equation}

Moreover, for the interpolation Geodesic, the following result holds.
\begin{proposition}[Interpolation between Gaussian distributions~\citep{mccann1997convexity} example 1.7]
Given the two Gaussian distributions defined in theorem (\ref{theorem:w_gaussian}), respectively. A constant speed geodesic is defined by the the path $(G^g_t)_{t \in [0, 1]}$, such that $\forall t \in [0, 1]$, and $G^g_t:= N(m_t, \Sigma_t)$ is given by 
\begin{equation} %\scriptsize
\begin{aligned}
& m_t = (1 - t) \mu_1 + t \mu_2, \\ 
&\Sigma_t = ((1 - t) Id + t T) \Sigma_1 ((1 - t) Id + t T)
\end{aligned}    
\end{equation}
where $T= \Sigma_1^{1/2}(\Sigma_1^{1/2}\Sigma_0\Sigma_1^{1/2})^{-1/2}\Sigma_1^{1/2}$.
\end{proposition}
We have illustrated that the Wasserstein barycenter, on the geodesic, of two Gaussian distributions is still a Gaussian distribution and admits an explicit form.
\begin{equation}
\begin{aligned}
\text{where, } & T^x_t(x) =(1 -t) x+ t T^x (x), \\
    & T^y_t(y) =(1 - t) y+ t T^y (y)
\end{aligned}
\end{equation}

\subsection{The effect of the regularization}

Recall that the geodesic regularization term is 
\begin{equation}
\begin{aligned}
\text{Reg}^{T^{xy}}_{l}(f_\theta) =
\int_{t=0}^1  \left|  \frac{d}{dt} \int_{\mathcal{X} \times \mathcal{Y}} l ( f_\theta(T^x_t(x), T^y_t(y))
    d P^0(x,y) \right|  dt 
\end{aligned}    
\end{equation}

Recall the classifier to be $f_\theta(x) = \langle \theta, x \rangle = \theta^\top x$, and the loss to be $l(f(x), y) = -y f(x)$, 

Consider the following minimization problem 
\begin{equation}
    \min_{f \in \mathcal{H}} \mathbb{E}_{(x,y) \sim P_{all}} [l(f(x), y)] + \frac{\lambda_1}{2} \text{Reg}^{T^{xy}}_{l}(f_\theta)
\end{equation}
Proof. The problem above can be written as follows
\begin{equation}
    \min_{\theta \in \mathbb{R}^d} \mathcal{L}(\theta) = - ( \langle \theta, \mu_1 - \mu_0 \rangle ) + \lambda_1 \text{Reg}(\theta)
\end{equation}

And under the above setting, we have 
\begin{equation}
\begin{aligned}
\text{Reg}^{T^{xy}}_{l}(f_\theta) &=
\int_{t=0}^1  \left|  \frac{d}{dt} \int_{\mathcal{X} \times \mathcal{Y}} l ( f_\theta(T^x_t(x), T^y_t(y))
    d P^0(x,y) \right|  dt \\
&= \int_{t=0}^1  \left|  \frac{d}{dt} \mathbb{E}_{x_0. y_0 \sim P^0(x,y)} [ l ( f_\theta(T^x_t(x), T^y_t(y)) ]
     \right|  dt  \\
&= \int_{t=0}^1  \left|  \frac{d}{dt} \mathbb{E}_{x_0. y_0 \sim P^0(x,y)} [ - T^y_t (y_0) \langle \theta, T^x_t(x_0) \rangle  ]
     \right|  dt  \\
     &\text{Recall the parameterization of the map as interpolation, as e.q., \ref{eq:adv_da_geo_interpolate}, }  \\
% &= \int_{t=0}^1  \left|  \frac{d}{dt} \mathbb{E}_{x_0. y_0 \sim P^0(x,y)} [ - [(1-t)y_0 + t \hat{T}^y (y_0)] \langle \theta, (1-t)x_0 + t \hat{T}^x(x_0) \rangle  ]
%      \right|  dt  \\
\end{aligned}    
\end{equation}

Here, using the dominated convergence theorem. Also, use the closed form interpolation $T^x_t(x_0) = (1-t)x_0 + t\hat{T}(x_0)$, and  $T^y_t(y_0) = 2t-1$ ($y_0 = -1$ and $y_1 = 1$), the above equation turns into
\begin{equation}
    \begin{aligned}
    &= \int_{t=0}^1  \left|  %\frac{d}{dt} 
    \mathbb{E}_{x_0. y_0 \sim P^0(x,y)} [
    -2 \langle \theta, T^x_t(x_0) \rangle - (2t-1) \langle \theta, \hat{T}(x_0) - x_0 \rangle  
    ] \right|  dt \\
    % & \text{The first term is the  expectation of geodesic and the second term is the W-distance.} \\
    % &= \int_{t=0}^1 \left| 
    % -2 (2t-1) \theta^\top \mu - 2(2t-1) \theta^\top \mu 
    % \right| dt
    \end{aligned}
\end{equation}
%\TBD{Introduce some variable from OT?} 
Since we have closed-form expression for the geodesic as $\mathbb{E}_{x_0 \sim P^0} [T^t(x_0)] = (2t-1)\mu$ and $\mathbb{E}_{x_0 \sim P^0}[\hat{T}(x_0) - x_0] = 2 \mu$, then
\begin{equation}
    \begin{aligned}
        &= \int_{t=0}^1 \left| 
    -2 (2t-1) \theta^\top \mu - 2(2t-1) \theta^\top \mu 
    \right| dt = \int_{t=0}^1 \left| 
    4(1-2t) \theta^\top \mu 
    \right| dt = 2 | \theta^\top \mu |
    \end{aligned}
\end{equation}

Then, the objective becomes 
\begin{equation}
    \min_\theta \mathcal{L}(\theta) := - \langle \theta, \mu \rangle + \frac{\lambda_1}{2} | \langle \theta, \mu \rangle |^2 + \frac{\lambda_2}{2} \| \theta\|^2_2 
    \label{eq:geo_reg_objective}
\end{equation}
Setting the first-order derivative to be zero, we obtain the solution
\begin{equation}
\begin{aligned}
    \nabla_\theta \mathcal{L}(\theta) &= - \mu + \lambda_1 \mu \otimes \mu \theta + \lambda_2 \theta = 0, \\
    \theta^* &= (\lambda_1 \mu \otimes \mu + \lambda_2 I_m)^{-1} \mu
\end{aligned}
\end{equation}
From Eq. (\ref{eq:geo_reg_objective}), we can see that geodesic regularization is a data-adaptive regularization that smooths the classifier according to the distribution of original data.

\subsection{ Proof of Proposition~\ref{prop:DRO-geodesic}}
\label{subsec:proof of prop DRO-Geodesic}
\paragraph{Assumptions and generalisation}
Without loss of generalisation, we assume $\sigma=1$ and $\mu=(1,0,\ldots,0)$, with $1$ at the first coordinate and $0$ elsewhere. Thus $\|\mu\|=1$.

Since $\eta(x|y=-1), \eta(x|y=1)$ are both Gaussian, assume $\eta(x|y=-1)\sim N(\psi_{-1},\Sigma_{-1})$ and $\eta(x|y=1)\sim N(\psi_1,\Sigma_1)$. 

Let $u_1,\ldots,u_d$ be the eigenvalues of $\Sigma_0$
Following Theorem~\ref{theorem:Wasserstein-Gaussian}, we have \begin{equation}\small
\begin{aligned}
W^2_2(N_d(\psi_1,\Sigma_1), N_d(\mu,\mathbb{I})) &=  \|\psi_1 - \mu \|^2 + tr(\Sigma_1) + d - 2 tr ({\Sigma_1}^{1/2}) \\
&= \|\psi_1 - \mu \|^2 +\sum_{i=1}^d\left(\sqrt{u_i}-1\right)^2
\end{aligned} 
\end{equation}

Similarly,
\begin{equation}\small
\begin{aligned}
W^2_2(N_d(\psi_{-1},\Sigma_{-1}), N_d(-\mu,\mathbb{I})) = \|\psi_{-1} + \mu \|^2 +\sum_{i=1}^d\left(\sqrt{v_i}-1\right)^2,
\end{aligned} 
\end{equation}
where $\{v_i\}_{1}^d$ are the eigenvalues of $\Sigma_{-1}$.

By linearity property of Gaussian distributions $\eta(X|Y=y) \sim N(\psi_y,\Sigma_y)$ implies, $\alpha'X+\beta|Y=y \sim N(\alpha'\psi_y + \beta,\alpha'\Sigma_y\alpha)$. Therefore, 
\begin{eqnarray}
    \mathbb{E}_{(X,Y) \sim \eta(X|Y)\nu(Y)}l(f_{(\alpha, \beta)}(X),Y)\propto P_{N(0,1)}\left(Z_1<-\dfrac{(\alpha'\psi_1 +\beta)}{\sqrt{\alpha'\Sigma_1\alpha}}\right)+P_{N(0,1)}\left(Z_{-1}>-\dfrac{(\alpha'\psi_{-1} +\beta)}{\sqrt{\alpha'\Sigma_{-1}\alpha}}\right), 
\end{eqnarray}
where $Z_1=\dfrac{\alpha'(X - \psi_1)}{\sqrt{\alpha'\Sigma_{1}\alpha}} \biggr|Y=1 \sim N(0,1)$ if $X|Y=1 \sim N(\psi_1,\Sigma_1)$ and \\$Z_{-1}=\dfrac{\alpha'(X - \psi_{-1})}{\sqrt{\alpha'\Sigma_{-1}\alpha}}\biggr|Y=-1 \sim N(0,1)$ if $X|Y=-1 \sim N(\psi_{-1},\Sigma_{-1})$. 

The LHS of Eq.~\eqref{eq: maxinf-gaussian-geodesic} becomes \begin{eqnarray}
\label{eq: maxinf-gaussian-geodesic_2}
    &  \max_{(\psi_1,\psi_{-1},\Sigma_1,\Sigma_{-1})\in B_{\epsilon}(\mu) } \inf_{\alpha, \beta}P_{N(0,1)}\left(Z_1<-\dfrac{(\alpha'\psi_1 +\beta)}{\sqrt{\alpha'\Sigma_1\alpha}}\right)+P_{N(0,1)}\left(Z_{-1}>-\dfrac{(\alpha'\psi_{-1} +\beta)}{\sqrt{\alpha'\Sigma_{-1}\alpha}}\right),
\end{eqnarray}
where, $B_{\epsilon}(\mu)=\{(\psi_1,\psi_{-1},\Sigma_1,\Sigma_{-1}): \|\psi_{-1} + \mu \|^2 +\sum_{i=1}^d\left(\sqrt{v_i}-1\right)^2 \leq \epsilon^2, \|\psi_{1} - \mu \|^2 +\sum_{i=1}^d\left(\sqrt{u_i}-1\right)^2 \leq \epsilon^2\}$
$\{u_i\}_1^d$ are eigenvalues of $\Sigma_{-1}$, and $\{u_i\}_1^d$ are eigenvalues of $\Sigma_{-1}$.
\paragraph{Step 1:}
First, we show the proof under the assumption $\Sigma_1=\Sigma_{-1}=\Sigma$.

Fixing $\psi_1,\psi_{-1}, \Sigma$ so that $\psi_1,\psi_{-1}, \Sigma, \Sigma \in B_{\epsilon}(\mu)$ and setting derivative to $0$ of 
\begin{eqnarray}
\label{eq: maxinf-gaussian-geodesic_3}
    P_{N(0,1)}\left(Z_1<-\dfrac{(\alpha'\psi_1 +\beta)}{\sqrt{\alpha'\Sigma\alpha}}\right)+P_{N(0,1)}\left(Z_{-1}>-\dfrac{(\alpha'\psi_{-1} +\beta)}{\sqrt{\alpha'\Sigma\alpha}}\right)
\end{eqnarray}
 w.r.t $\beta$ yields either (i) $\alpha'\psi_1 +\beta=\alpha'\psi_{-1} +\beta$, or (ii) $\alpha'\psi_1 +\beta=-(\alpha'\psi_{-1} +\beta)$.

Setting derivative of \eqref{eq: maxinf-gaussian-geodesic_3} w.r.t $\alpha$ to $0$ yields $\psi_1=\psi_{-1}$, which clearly is a contradiction for $\epsilon<1$.

Considering (ii) above yields $\beta=-\alpha'(\psi_{-1} + \psi_1 )/2$. Then Eq.~\eqref{eq: maxinf-gaussian-geodesic_3} leads to 
\begin{eqnarray}
\label{eq: maxinf-gaussian-geodesic_4}
    &  P_{N(0,1)}\left(Z>\dfrac{\alpha'(\psi_1 -\psi_{-1})}{2\sqrt{\alpha'\Sigma\alpha}}\right)+P_{N(0,1)}\left(Z<-\dfrac{\alpha'(\psi_1 -\psi_{-1})}{2\sqrt{\alpha'\Sigma\alpha}}\right) \nonumber\\
    &=2P_{N(0,1)}\left(Z>\dfrac{\alpha'(\psi_1 -\psi_{-1})}{2\sqrt{\alpha'\Sigma\alpha}}\right).
\end{eqnarray}
The RHS above is minimized w.r.t $\alpha$ when $\dfrac{\alpha'(\psi_1 -\psi_{-1})}{2\sqrt{\alpha'\Sigma\alpha}}$ is maximized. A simple calculation shows that this is achieved for $\alpha=\dfrac{\tilde{a}\Sigma^{-1}}{\sqrt{\tilde{a}'\Sigma^{-1}\tilde{a}}}$, where $\tilde{a}= (\psi_1 -\psi_{-1})$. This gives $\dfrac{\alpha'(\psi_1 -\psi_{-1})}{\sqrt{\alpha'\Sigma\alpha}}= {\sqrt{\tilde{a}'\Sigma^{-1}\tilde{a}}}$.

For a given $\Sigma$, this is minimized (and equivalently Eq.~\eqref{eq: maxinf-gaussian-geodesic_3} is maximised) when $ (\psi_1 -\psi_{-1})/2$ is in the direction of the largest eigenvector of $\Sigma$, say $v_1$, and $\tilde{a}'\Sigma^{-1}\tilde{a}=\|\psi_1 -\psi_{-1}\|^2 /v_1$.

Notice that since the other eigenvalues do not affect the choice of $(\psi_1 -\psi_{-1})$, based on the constraints $\|\psi_{-1} + \mu \|^2 +\sum_{i=1}^d\left(\sqrt{v_i}-1\right)^2 \leq \epsilon^2$ and $\|\psi_{1} + \mu \|^2 +\sum_{i=1}^d\left(\sqrt{v_i}-1\right)^2 \leq \epsilon^2$, where $v_i$ are the ordered eigenvalues of $\Sigma$, we can assume all other eigenvalues to be $1$ without affecting the value in the RHS of Eq.~\eqref{eq: maxinf-gaussian-geodesic_4}. 

Let the largest eigenvalue of $\Sigma$ be $(1+\delta)^2$. Then based on the constraints $\|\psi_1 -\psi_{-1}\|^2 /v_1$ is minimized when $\psi_1=-\psi_{-1}$ and they are concurrent with $\mu$ and $-\mu$. 

Based on the constraints, this yields,

\begin{eqnarray}
\sqrt{\|\psi_1 -\psi_{-1}\|^2 /v_1}= 2(1-\sqrt{\epsilon^2-\delta^2})/(1+\delta),
\end{eqnarray}
 which is an increasing function of $\delta$ and is minimised at $\delta=1$, which implies $\Sigma=\mathbb{I}$, and $\psi_1=(1-\epsilon)\mu$, $\psi_{-1}=(1-\epsilon)-\mu$, which are the geodesic optimizers. 

 Therefore, 
 \begin{eqnarray}
    & \max_{\tilde{A}(\nu_{\mu},\epsilon)} \inf_{\alpha, \beta}  \mathbb{E}_{(X,Y) \sim \eta(X|Y)\nu(Y)}l(f_{(\alpha, \beta)}(X),Y) = \nonumber \\
    & \mathbb{E}_{(X,Y) \sim \nu_{(1-\epsilon)\mu}(X|Y)\nu(Y)}l(f_{(\mu, 0)}(X),Y).
\end{eqnarray}
where $\tilde{A}(\nu_{\mu},\epsilon)= \{\eta(X|Y): \eta(\cdot|Y) $ is Gaussian, $ W(\eta(X|Y),\nu_{\mu}(X|Y)) \leq \epsilon, \ \forall Y\in \{-1,1\}$, $Cov(\eta(\cdot|Y=1))=Cov(\eta(\cdot|Y=-1))\}$

\paragraph{Step 2:} Next, we use the result in Step 1 to show the remaining.

Therefore, consider the situation where we do not restrict to $\Sigma_1=\Sigma_{-1}$. 

Set $\Sigma^0_1=\Sigma^0_{-1}=\mathbb{I}$, $\psi^0_1=(1-\epsilon)\mu$, $\psi^0_{-1}=-(1-\epsilon)\mu$, $\alpha_0=(1-\epsilon)\mu$, $\beta_0=0$. Also, let 
\begin{eqnarray}
    F(\psi_1,\psi_{-1},\Sigma_1,\Sigma_{-1},\alpha,\beta):=\mathbb{E}_{(X,Y) \sim \eta(X|Y)\nu(Y)}l(f_{(\alpha, \beta)}(X),Y),
\end{eqnarray}
where $\eta(X|Y=1)\sim N(\psi_1,\Sigma_1)$, $\eta(X|Y=-1)\sim N(\psi_{-1},\Sigma_{-1})$.

Note that $F(\psi^0_1,\psi^0_{-1},\Sigma^0_1,\Sigma^0_{-1},\alpha_0,\beta_0)={E}_{(X,Y) \sim \nu_{(1-\epsilon)\mu}(X|Y)\nu(Y)}l(f_{(\mu, 0)}(X),Y)$.

From \textbf{Step 1 } of the proof, we have shown that,  $F(\psi^0_1,\psi^0_{-1},\Sigma^0_1,\Sigma^0_{-1},\alpha,\beta) \geq F(\psi^0_1,\psi^0_{-1},\Sigma^0_1,\Sigma^0_{-1},\alpha_0,\beta_0)$.

Now, we claim it is enough to show $F(\psi_1,\psi_{-1},\Sigma_1,\Sigma_{-1},\alpha_0,\beta_0) \leq F(\psi^0_1,\psi^0_{-1},\Sigma^0_1,\Sigma^0_{-1},\alpha_0,\beta_0)$ for $\psi_1,\psi_{-1},\Sigma_1,\Sigma_{-1} \in {A}(\nu_{\mu},\epsilon)$, since then 

\begin{eqnarray}
     & &\max_{\tilde{A}(\nu_{\mu},\epsilon)} inf_{\alpha,\beta}F(\psi^0_1,\psi^0_{-1},\Sigma^0_1,\Sigma^0_{-1},\alpha,\beta) \nonumber \\ &\geq &inf_{\alpha,\beta}F(\psi^0_1,\psi^0_{-1},\Sigma^0_1,\Sigma^0_{-1},\alpha,\beta) \nonumber \\ &\geq &F(\psi^0_1,\psi^0_{-1},\Sigma^0_1,\Sigma^0_{-1},\alpha_0,\beta_0)
     \nonumber \\ &\geq &\max_{\tilde{A}(\nu_{\mu},\epsilon)} F(\psi_1,\psi_{-1},\Sigma_1,\Sigma_{-1},\alpha_0,\beta_0) \nonumber \\ &\geq &
       inf_{\alpha,\beta} \max_{\tilde{A}(\nu_{\mu},\epsilon)} F(\psi^0_1,\psi^0_{-1},\Sigma^0_1,\Sigma^0_{-1},\alpha,\beta)  
\end{eqnarray}
and we know that 
\begin{eqnarray}
    inf_{\alpha,\beta} \max_{A(\nu_{\mu},\epsilon)} F(\psi^0_1,\psi^0_{-1},\Sigma^0_1,\Sigma^0_{-1},\alpha,\beta) \geq \max_{A(\nu_{\mu},\epsilon)} inf_{\alpha,\beta}F(\psi^0_1,\psi^0_{-1},\Sigma^0_1,\Sigma^0_{-1},\alpha,\beta)
\end{eqnarray}

This would imply that 
\begin{eqnarray}
    inf_{\alpha,\beta} \max_{A(\nu_{\mu},\epsilon)} F(\psi^0_1,\psi^0_{-1},\Sigma^0_1,\Sigma^0_{-1},\alpha,\beta) &=& \max_{A(\nu_{\mu},\epsilon)} inf_{\alpha,\beta}F(\psi^0_1,\psi^0_{-1},\Sigma^0_1,\Sigma^0_{-1},\alpha,\beta) \nonumber \\
    &=& {E}_{(X,Y) \sim \nu_{(1-\epsilon)\mu}(X|Y)\nu(Y)}l(f_{(\mu, 0)}(X),Y)
\end{eqnarray}

and complete the proof.

Now, 
\begin{eqnarray}
F(\psi_1,\psi_{-1},\Sigma_1,\Sigma_{-1},\alpha_0,\beta_0)=P\left(Z<\dfrac{\mu'\psi_{-1}}{\sqrt{\mu'\Sigma_{-1}\mu}}\right) +P\left(Z> \dfrac{\mu'\psi_1}{\sqrt{\mu'\Sigma_{-1}\mu}}\right).
\end{eqnarray}

Since the constraints in $A(\nu_{\mu},\epsilon)$ are separate and symmetric with respect to $\psi_1,\Sigma_{1}$ and $\psi_{-1},\Sigma_{-1}$ respectively, and standard Normal distribution is symmetric. Therefore the max of $F(\psi_1,\psi_{-1},\Sigma_1,\Sigma_{-1},\alpha_0,\beta_0)$ over $A(\nu_{\mu},\epsilon)$ is attained when $\psi_{-1}=-\psi_1$ and $\Sigma_1=\Sigma_{-1}$. The remainder follows from \textbf{Step 1}.

\subsection{Error probabilities in closed form (Definition~\ref{definition:classification_error_prob})}

\paragraph{Standard accuracy probability.} The standard accuracy probability is given as 
\begin{equation}   
    \mathbb{P}_{x \sim D} (f_\theta(x) = y) = 1 - \mathbb{P}_{x \sim D} (y x^\top \theta < 0) = 1 - Q\left(\frac{\mu^\top \theta}{ \sigma \| \theta \|} \right)
\end{equation}
where $Q(x) = \frac{1}{\sqrt{2 \pi}} \int^\infty_x e^{- t^2 / 2} dt$.

\paragraph{$l_\infty^\epsilon$ robust accuracy.}
The $l_{\infty}^\epsilon$ robust accuracy is given as:

\begin{align}
\inf_{\|v \|_\infty < \epsilon} \mathbb{P}_{x \sim D} (f_\theta(x) = y) &= 1 - \mathbb{P}_{x \sim D}\left( \inf_{\|v \|_\infty < \epsilon} \left\{ y \cdot (x + v)^\top \theta \right\} <0 \right) \nonumber \\
    &= 1 - Q\left( \frac{\mu^\top \theta}{\sigma \| \theta \|} - \frac{\epsilon \| \theta \|_1}{\sigma \| \theta \|} \right)
\end{align}    

\paragraph{$l_2^{\sigma_s}$ smoothed classifier for certifiable robust accuracy.} It is the accuracy probability from a $N(0, \sigma_s^2 I)$ smoothed classifier:
\begin{align}
    \mathbb{P}_{x \sim D, \delta \sim N(0, \sigma_s^2 I)} (f_\theta(x + \delta) = y)
\end{align}

The above probability involves two random variables $x(y) \sim D$ and $\delta \sim N(0, \sigma_s^2 I)$, which are independent of each other. This addition is considered as the convolution of two pdfs. Since $D$ is a mixture of Gaussian and $\delta$ subjects to a Gaussian, there is a closed form 
\begin{align}
    D^* = N(\mu, \sigma^2 I) \ast N(0, \sigma_s^2 I) = N(\mu, \sigma^2 + \sigma_s^2 I)
\end{align}
Thus, the certifiable robust accuracy in this setting is equivalent to the standard accuracy probability from a convoluted dataset.
\begin{align}
    \mathbb{P}_{x \sim D^*} (f_\theta(x) = y) =  1 - Q\left(\frac{\mu^\top \theta}{ (\sigma + \sigma_s) \| \theta \|} \right)
\end{align}

%\subsection{Proof of Theorem 1}

\subsection{Provable improvement from data augmentation (Theorem~\ref{theorem:Data_aug_Gaussian})}

\begin{proof}
We divide the proof into several steps as follows:

\paragraph{Step 1:} The distribution of original data $(X,Y)$ is iid with $X|Y \sim N (y\mu, \sigma^2 I)$. Thus, from Eq.(10) and (11) in~\cite{carmon2019unlabeled_imp_rob}, we have
\begin{align}
\mathbb{PE}^{p, \epsilon}_a(f_\theta) = Q\left(\dfrac{\mu^T\theta}{\sigma \|\theta\|} - \dfrac{\epsilon\sqrt{d}}{\sigma}\right), ~\text{where} ~
Q(x)= \dfrac{1}{2\pi}\int_x^{\infty} e^{-z^2/2} \mathrm{d}z.
\end{align}
Let $\alpha(n_0,n_1)$ be used to denote $n_1/(n_0+n_1)$. Additionally, let $\hat{{\theta}}_{n_1}=\dfrac{\left(\sum_{i=1}^{n_1}\tilde{Y}_i \tilde{X}_i \right)}{n_1 }$.

\paragraph{Step 2:}
Note that 
$
\mathbb{PE}^{p, \epsilon}_a(f_{\hat{\theta}_{n_0}}) \geq   \mathbb{PE}^{p, \epsilon}_a(f_{\hat{\theta}_{n_0,n_1}})
$
holds as long as $\dfrac{\mu^T\hat{\theta}_{n_0}}{\|\mu\| \|\hat{\theta}_{n_0}\|} \leq \dfrac{\mu^T\hat{\theta}_{n_0,n_1}}{\|\mu\| \|\hat{\theta}_{n_0,n_1}\|}$. This is what we will show.

\paragraph{Step 3:} 
Let $\hat{\theta}_{n_1}\in \{x: \|x-t\mu\| \leq A_3:= t \log(n_1)/n_1\} $.
Also assume:
\begin{align}
& \hat{\theta}_{n_0}\in \{x:A_1:=(t+n_0)/\sqrt{n_1} \leq \|x-\mu\|  A_2:=\log(n_1)\} \cap \{x:  \mu^T(x) /\|x\| \\
& \leq \sqrt{\|\mu\|^2- ((t+n_0)/\sqrt{n_1})^2\} \}}
\end{align}

%$\hat{\theta}_{n_0}\in \{x:A_1:=(t+n_0)/\sqrt{n_1} \leq \|x-\mu\| \leq A_2:=\log(n_1)\} \cap \{x:  \mu^T(x) /\|x\| \leq \sqrt{\|\mu\|^2- ((t+n_0)/\sqrt{n_1})^2\} \}}$. 

We show that in this case  $\dfrac{\mu^T\hat{\theta}_{n_0}}{\|\mu\| \|\hat{\theta}_{n_0}\|} \leq \dfrac{\mu^T\hat{\theta}_{n_0,n_1}}{\|\mu\| \|\hat{\theta}_{n_0,n_1}\|}$. Indeed, 
\begin{eqnarray}\small
\dfrac{\mu^T\hat{\theta}_{n_0}}{\|\mu\| \|\hat{\theta}_{n_0}\|} 
&\leq& \cos(\pi/2-\arcsin (A_1/\|\mu\|)) \nonumber \\ 
&<& \cos\left(\pi_2-\arcsin\left(\dfrac{n_1A_3 +n_0A_2}{n_1t\|\mu\|+n_0\|\mu\|}\right)\right) \nonumber\\ 
&<& \dfrac{(n_1t\mu+n_0\mu)^T\hat{\theta}_{n_0,n_1}}{\|n_1t\mu+n_0\mu\| \|\hat{\theta}_{n_0,n_1}\|} \\
&=& \dfrac{\mu^T\hat{\theta}_{n_0,n_1}}{\|\mu\| \|\hat{\theta}_{n_0,n_1}\|}.
\end{eqnarray}

The first inequality holds by Cauchy-Schwartz, while the second inequality holds by substituting values of $A_1,A_2$ and $A_3$ (note that  $n_1A_3+n_0A_2 \leq (n_1t +n_0)A_1$) combined with the facts that $tN_1 +n_0 >\sqrt{N_1} \log(N_1)$ and $f(n)=tn +n_0 -\sqrt{n} \log(n)$ is monotonically increasing in $n$.

Note that $\hat{\theta}_{n_0,n_1}=\alpha(n_0,n_1)\hat{\theta}_{n_1}+(1-\alpha(n_0,n_1))\hat{\theta}_{n_0}$. 
Moreover, $\hat{\theta}_{n_1}$ and $\hat{\theta}_{n_0} $ follow $\hat{\theta}_{n_1} \sim N(t\mu,\sigma^2 I_d/n_1)$ and  $\hat{\theta}_{n_0} \sim N(\mu,\sigma^2 I_d/n_0)$, independently.

The third inequality therefore holds because $\|y-t\mu\|\leq A_3$ and $\|x-\mu\| \leq A_2$ implies 
$\|\alpha(n_0,n_1)y+(1-\alpha(n_0,n_1))x- \alpha(n_0,n_1)t\mu+(1-\alpha(n_0,n_1))\mu\| \leq \alpha(n_0,n_1)A_3+(1-\alpha(n_0,n_1))A_2$. 

Therefore, $P\left(\dfrac{\mu^T\hat{\theta}_{n_0}}{\|\mu\| \|\hat{\theta}_{n_0}\|} \leq \dfrac{\mu^T\hat{\theta}_{n_0,n_1}}{\|\mu\| \|\hat{\theta}_{n_0,n_1}\|}\right) \geq P(\hat{\theta}_{n_1}\in \{x: \|x-t\mu\| \leq A_3:= t \log(n_1)/n_1\}) \times P(\hat{\theta}_{n_0}\in \{x:A_1:=(t+n_0)/\sqrt{n_1} \leq \|x-\mu\| \leq A_2:=\log(n_1)\} \cap \{x:  \mu^T(x) /\|x\| \leq \sqrt{\|\mu\|^2- ((t+n_0)/\sqrt{n_1})^2\} \}})$. 

The result now holds true by simple variable transformations.
\end{proof}

\section{Additional intuition \& related works}
\subsection{Related works}

Nonparametric gradient flow~\citep{liutkus2019sliced} provides satisfactory data generation with theoretical guarantees. Further, data synthesis~\citep{hua2023dynamic} on the feature-Gaussian manifold can be realized by the  maximum mean discrepancy (MMD) gradient flow~\citep{mroueh2021convergence} is proven to be effective for transfer learning tasks.

Recent studies have shown that data augmentation, including potentially unlabelled data, improves adversarial robustness, both empirically and theoretically. 
A line of study is rigorously explaining how the adversarial robustness is affected by the sample size \cite{schmidt2018adversarially} where unlabeled or out-of-distribution data are helpful~\citep{carmon2019unlabeled_imp_rob,deng2021improving_ood_data_rob_gaussian_mix,dan2020sharp_rob_gaussian,bhagoji2019lower_adv_ot}.
Although conventional data augmentation techniques (such as random flips, rotations, or clips)~\citep{howard2013some_data_aug,he2015convolutional_data_aug} have remarkable standard learning performance, is not until recently that researchers started to investigate how data augmentation can improve deep learning robustness~\citep{volpi2018generalizing_data_aug_dro,Ng2020SSMBASM,rebuffi2021data_aug_rob_deepmind}. 

A few studies have already explored the usage of optimal transport (OT) ideas within mixup. PuzzelMix~\citep{kim2020_puzzlemix} aligns saliency regions of images with a masked optimal transport. 
The idea of augmentation with saliency is extended in recent works~\citep{kim2021_co-mixup}.  
OT also helps align feature tensors in high-dimensional vector space~\citep{venkataramanan2022alignmix}, while this method also requires autoencoder models to capture reasonable representations, they perform barycentric projection (linear map) and implicitly restrict the interpolation manifold. 
The idea of barycenter is also used in AutoMix~\citep{zhu2020_automix_optransmix}, which relies on a barycenter generator. On the other hand, OT has been shown to be effective in various areas such as data augmentation for cardiovascular diseases~\citep{zhu2022geoecg,qiu2022optimal}, model personalization~\citep{zhu2022physiomtl}, and multimodal learning~\cite{qiu2022semantics}. 

% \TBD{Less on certifiable robustness.. }
% The recently proposed idea of randomized smoothing~\citep{pmlr-v97-cohen19c} provides certifiably robust classifier on $\ell_2$-perturbations. This notion is desirable as it not only offers a provable guarantee of the robustness of deep neural networks but also can be obtained in a relatively computational-efficient fashion. To improve the robustness, multiple approaches have been proposed to facilitate the training of base classifiers that could have better certified robustness. In addition to different regularization methods~\citep{Zhai2020MACER,jeong2020consistency,li2018certified_renyi}, the pioneer work adopted Gaussian data augmentation~\citep{pmlr-v97-cohen19c} to pursue smoothness, augmenting more complicated smoothing distribution~\citep{li2021tss_TSS}, adversarial samples~\citep{salman2019provably}, and mixing adversarial samples with original samples~\citep{jeong2021smoothmix} are all promising methods.

\subsection{Intuitions}
The recently proposed idea of randomized smoothing~\citep{cohen2019certified_cert_rob} provides a certifiably robust classifier on $\ell_2$-perturbations. This notion is desirable as it not only offers a provable guarantee of the robustness of deep neural networks but also can be obtained in a relatively computationally-efficient fashion. To improve the robustness, multiple approaches have been proposed to facilitate the training of base classifiers that could have better certified robustness. In addition to different regularization methods~\citep{Zhai2020MACER,jeong2020consistency,li2018certified_renyi}, the pioneering work adopted Gaussian data augmentation~\citep{cohen2019certified_cert_rob} to pursue smoothness, augmenting more complicated smoothing distribution~\citep{li2021tss_TSS}, adversarial samples~\citep{salman2019provably}, and mixing adversarial samples with original samples~\citep{jeong2021smoothmix} are all promising methods.

We can illustrate our intuitions as follows:
{(1)} Instead of the datapoint-specific adversarial perturbations that are aimed to attack one specific sample, the directed augmented data distribution can be considered as universal perturbations~\citep{moosavi2019robustness} that cause label change for a set of samples from the perturbed distribution $\mathcal{U}_P$.
{(2)} Such perturbation matches the global manifold structure of the dataset~\citep{greenewald2021_kmixup}, therefore promoting a smoother decision boundary.
{(3)} 
It is shown in~\cite{wei2020theoretical_expansion} that this augmentation strategy improves the expansion of the neighborhood of class-conditional distributions.  
This formulation allows us to employ the results from OT theories~\citep{villani2009optimal} and Wasserstein Barycenter~\citep{agueh2011barycenters} thus firmly estimating the perturbed distribution $\mathcal{U}_P$.
{(4)} Apart from most data augmentation techniques that improve the generalization by creating samples that are likely to cover the testing distribution. Our argumentation promotes the inter-class behavior~\citep{tokozume2017learning_between_class} and potentially enlarges the ratio of the between-class distance to the within-class variance~\citep{fisher1936use}, or Fisher's criterion.

% \begin{definition}
% (Geodesics in Wasserstein space). Let $\mu$ and $\nu$ be two distributions. Consider a map $m: [0, 1] \mapsto \mathcal{M}(\mathcal{X})$ taking $[0, 1]$ to the set of distributions, such that $m(0) = \mu$ and $m(1) = \nu$. 
% Thus $(p_{\alpha}: 0 \leq \alpha \leq 1)$ is a path connecting $\mu$ and $\nu$, where $p_{\alpha}=m(\alpha)$. The length of $m$ --- denoted by $L(m)$ --- is the supremum of $\sum_{i=1}^{K} W\left(m\left(\alpha_{i-1}\right), m\left(\alpha_{i}\right)\right)$ over all $m$ and all $0=\alpha_{1}<\cdots<\alpha_{K}=1$. 
% Therefore, there exists such a path $m$ such that $L(m) = W(\mu, \nu)$ and $(p_{\alpha}: 0 \leq \alpha \leq 1)$ is the geodesic connecting $\mu$ and $\nu$. 
% \end{definition}

% \begin{definition}
%  (Wasserstein Barycenter). The Wasserstein barycenter of a set of measures $\{\nu_1, ..., \nu_N \}$ in a probability space $\mathbb{P} \subset \mathcal{M}(\mathcal{X})$ is a minimizer of objective $f_{wb}$ over $\mathbb{P}$, where
% $
% f_{wb}(\mu) := \frac{1}{N} \sum_{i=1}^N \alpha_i W(\mu, \nu_i),
% $
% %\begin{equation}\small
% %     f_{wb}(\mu) := \frac{1}{N} \sum_{i=1}^N \alpha_i W(\mu, \nu_i),
% %\end{equation}
% and $\alpha_i$ are the weights such that $\sum \alpha_i = 1$ and $\alpha_i > 0$. 
% \end{definition}

\subsection{Mixture of distributions}

\paragraph{Mixture of distributions in classification}
While it is challenging to specify a general distribution family for $\mathcal{U}_P$ to enable robust training, we further look into the structure of the joint data distribution $\mathcal{P}^{all}_{x,y}$.
Considering a $k$-class classification problem, it is natural to view the data distribution as a mixture of subpopulation distributions 
\begin{align}\small
\mathcal{P}^{all}_{x,y} = \sum_{i=1}^k w_k P_k(X, Y), 
\end{align}
where each mixture $P_k(X, Y)$ stands for an individual class~\citep{carmon2019unlabeled_imp_rob,zhai2019adversarially_rob_gaussian,dan2020sharp_rob_gaussian}.
In such a case, we can utilize the geometric structure underneath the training data to specify the adversarial distribution and improve the model's robustness, especially when differentiating subpopulation distributions.

\subsection{Connection to Mixup }

Let us define $x \in \mathcal{X}$ to be input data and $y \in \mathcal{Y}$ to be the output label. Let  $(X_0,Y_0)\sim P^0_{x,y}$ and $(X_1,Y_1)\sim P^1_{x,y}$ be two distributions on $\mathcal{X} \times \mathcal{Y}$. We may also allow $Y_1=1$ a.s. $P^1_{x,y}$ and $Y_0=0$ a.s. $P^0_{x,y}$.For brevity, we use $P^i_{x,y}$ and $P^i$ interchangeably. For the mixup setting~\citep{kim2020_puzzlemix}, the goal is to optimize a loss function as follows:
 \begin{eqnarray}
 \min_{f} \mathbb{E}_{(X_0,Y_0) \sim P^0_{x,y},(X_1,Y_1)\sim P^1_{x,y}} \mathbb{E}_{ \lambda\sim q}  (l(f(h_{\lambda}(X_0,X_1)),e_{\lambda}(Y_0,Y_1)))
 \end{eqnarray}
where the label mixup function is $e_{\lambda}(y_0,y_1) = (1-\lambda)y_0 + \lambda y_1$. Input mixup uses $h_{\lambda}(x_0,x_1) = (1-\lambda)x_0+\lambda x_1$, and $q$ is the distribution of $\lambda$ considered.
Here conventional mixup~\citep{zhang2018mixup} considers the linear interpolation of independently sampled data. However, we feel this may be too restrictive in nature, since this may lead to the creation of samples that may not contribute in  situations when the task is binary classification. In that respect, consider the following minimization problem instead,
  \begin{eqnarray}
  \min_f  \mathbb{E}_{ t\sim q} \mathbb{E}_{(X_0,X_1,Y_0,Y_1) \sim \tilde{\pi} } \nonumber  \left[ l(f(h_{t}(X_0,X_1)), e_{t}(Y_0,Y_1) \right],
 \end{eqnarray}
 % \TBD{The benefit of such setting. Have optimal structure on specific local pair. N}
 where $\tilde{\pi}(x_0,x_1,y_0,y_1)=$ $\tilde{\pi}_x((x_0,x_1)|(y_0,y_1)) \times \tilde{\pi}_y((y_0,y_1))$, with $\tilde{\pi}_x((\cdot,*)|(y_0,y_1)) $ being the optimal transport coupling between the conditional distributions $P^0(\cdot|y_0)$ and $P^1(\cdot|y_1)$, while $\tilde{\pi}_y((\cdot,\cdot))$ is the optimal transport coupling between the marginal distributions of $Y_0$ and $Y_1$. It is easy to see that under the existence of unique Monge maps in-between marginal and conditional distributions of $Y_0, Y_1$ and $X_0|Y_0,X_1|Y_1$ respectively, this is equivalent to solving the following optimization problem.
\begin{eqnarray}
\label{eq:connect}
\min_f \mathbb{E}_{ t\sim q} \mathbb{E}_{(x,y)  \sim \mu^{\{P_0,P_1\}}_{t} }\left[ l(f(x), y) \right] 
\end{eqnarray}
where  Eq.\eqref{eq:barycenter} $\mu^{\{P_0,P_1\}}_{t} = \min_\mu U^{N=2}_{wb}(\mu) := \min_\mu (1-t) W (\mu, P^0) + tW(\mu, P^1)$  and $W(\cdot,*)$ is the Wasserstein metric.

\subsection{A more general problem formulation: multi-marginal optimal transport}
We restrict the adversarial distribution family to be the geodesic interpolations between individual subpopulation distributions. Thus eq.(\ref{eq:eq1_dro}) becomes 
\begin{align}
    \min_{f} \max_\alpha \mathbb{E}_{x, y \sim U^{wb}(\alpha) } \left[ l(f(x), y) \right] 
    \label{eq:adv_w_bc}
\end{align}
where $U^{wb}(\alpha )= \arg \min_U \frac{1}{K} \sum_i \alpha_i W (U, P_i)$ is the Wasserstein barycenter, in other words, the interpolation of subpopulation distributions.

In addition, with a predictive function $f$, we can consider a dynamic metric, \textit{geodesic loss}, $\text{R}^{\text{geo}}_f$ that measures the change of its performance with the criteria function $l$ while \textit{gradually interpolating} among subpopulation distributions $\{P_i\}_{i=1}^K$. This metric is thus a continuous function of $\alpha$, where
\begin{equation}\small
\text{R}^{\text{geo}}_f (\alpha) =  \mathbb{E}_{x, y \sim U^{wb}(\alpha)} \left[l ( f(x), y)
\right] 
\end{equation}
The geodesic loss $\text{R}^{\text{geo}}_f$ provides us a new lens through which we can measure, interpret, and improve a predictive model's robustness from a geometric perspective.

\section{Algorithm \& Computation }

%\begin{minipage}{0.66\textwidth}\small
\begin{algorithm}[H]\small
    \centering
    \caption{Sinkhorn Barycenter}\label{alg:sinkhorn}
    \begin{algorithmic}[1]
    \STATE \textbf{Input}: Empirical distributions $\alpha_1,\alpha_2$, cost matrix $\mathbf{C}$, $\mathbf{K} = exp (- \mathbf{C} / \epsilon)$.
    \STATE \textbf{Output}: debiased barycenters $\alpha_{S_\epsilon}$
    \STATE Initialize scalings $(b_1, b_2) ,d$ to $\mathbf{1}$.
    \WHILE{not converge} 
    \FOR{$k=1$ to $2$}
    \STATE $a_k \xleftarrow{} \left( \alpha_k / \mathbf{K} n_k \right)$
    \ENDFOR
    \STATE $\alpha  \xleftarrow{} d \odot \sum_{k=1}^K (\mathbf{K}^\top a_k)^{w_k}$
    \FOR{$k=1$ to $2$}
    \STATE $b_k  \xleftarrow{} \left( \alpha / \mathbf{K}^\top \alpha_k \right)$ 
    \ENDFOR
    \STATE $d \xleftarrow{} \sqrt{ d \odot (\alpha \mathbf{K} d)}$
    \ENDWHILE
    \end{algorithmic} 
\end{algorithm}
%\end{minipage}

In practice, we only observe discrete training samples that represent an empirical distribution of $P_i$ and $P_j$. 
Consider  $\mathbf{X}_i = \{\mathbf{x}^i_l\}_{l=1}^{n_i}$ and $\mathbf{X}_j = \{\mathbf{x}^j_l\}_{l=1}^{n_j}$ are two set of features from class $i$  and $j$ respectively. The empirical distributions are written as $\hat{P}_i = \sum_{l=1}^{n_i} p_l^i \delta_{x^i_l}$ and $\hat{P}_j = \sum_{l=1}^{n_j} p_l^j \delta_{x^j_l}$ where $\delta_{x}$ is the Dirac function at location $x \in \Omega$, $p_l^i$ and $p_l^j$ are probability mass associated to the sample. 
Then the Wasserstein distance, between empirical measures $\hat{P}_i$ and $\hat{P}_j$ becomes
\begin{equation}\small
    W(\hat{P}_i, \hat{P}_j)  = \inf_{\pi \in \hat{\Pi}} \sum_{l=1,k=1}^{n_i, n_j} c(\mathbf{x}^i_l, \mathbf{x}^j_k)  \pi_{l,k} + H(\pi),
\end{equation}
where $\hat{\Pi}:= \{\pi \in (\mathbb{R}^+)^{n_i \times n_j} | \pi \mathbf{1}_{n_j} = \mathbf{1}_{n_i} /n_i, \pi^\top  \mathbf{1}_{n_i} = \mathbf{1}_{n_j} /n_j \}$ with $\mathbf{1}_n$ a length $n$ vector of ones, $H(\cdot)$ is the negative entropy regularizer for us to utilize the Sinkhorn algorithm. $c(x, y)$ is the ground cost function that specifies the actual cost to transport the mass, or probability measure, from position $x$ to $y$.  Most studies merely use $l_2$ norm as the ground metric as there are a lot of desirable properties, such as the linear barycentric projection~\citep{villani2009optimal} used in other OT-based mixup methods~\cite{venkataramanan2022alignmix}.

\paragraph{Computation concerns: batch OT and entropic OT} Discrete optimal transport involves a linear program that has an $O(n^3)$ complexity. 
Hence, the potential computation issues can not be ignored. 

First of all, we adopted the celebrated entropic optimal transport~\citep{cuturi2013sinkhorn} and used the Sinkhorn algorithm to solve for OT objectives and Barycenters~\citep{janati2020debiased_barycenter} (algorithm 2). The Sinkhorn algorithm has a $O(n \log n)$ complexity, thus it can ease the computation burden. In addition, the pairwise Wasserstein distance can be precomputed and stored. Last but not least, we follow the concept of minibatch optimal transport~\citep{fatras2021minibatch_OT} where we sample a batch of samples from each condition during the data augmentation procedure. Whereas minibatch OT could lead to non-optimal couplings, our experimental results have demonstrated that our data augmentation is still satisfactory.  

\paragraph{Data augmentation }
In our work, we focus on studying the advantages and limitations brought by certain data augmentation algorithm $\mathcal{A}: \mathcal{X} \mapsto \mathcal{X}$ such that $\mathcal{A}_\# P^0_{X, Y} = P^1_{X, Y}$, where $A_\# \mu = \nu$ denotes that $\nu$ is the pushfoward measure of $\mu$ by $A$. 
%\TBD{Rewrite the data augmentation with our formulation.}
Under this notation, a data augmentation algorithm is a transport map that transforms the original data distribution $P^0_{X, Y}$ towards the augmentation distribution $P^1_{X, Y}$. In practice, the estimation of such a transport map is challenging. Nevertheless, we can assume the access to the distribution $P^1_{X,Y}$ and augmented data samples $\{ \Tilde{x}_i, \Tilde{y}_i \}_{i=1}^{n_1} \sim P^1_{X,Y}$. 
Consider a supervised learning algorithm $T(\cdot, \cdot): (\mathcal{X} \times \mathcal{X}) \mapsto \Theta$ that maps a dataset to a model parameter $\theta$. The standard training process relies on the raw dataset $\hat{\theta}_{n_0} = T(X, Y)$ while data augmentation provides additional dataset and  $\Tilde{\theta}_{n_0 + n_1} = T([X, Y], [\Tilde{X}, \Tilde{Y}])$.

%%%%%%%%%%%%%%%%%%%%%%%%%%%%%
%%%%%% Show data augmentation works
% \begin{theorem}
% Suppose the original data $\{ {X}_i, {Y}_i, \}_{i=1}^{{n_0}}$ satisfies $X|Y=y \sim N(y\mu, \sigma^2 I)$, $Y\sim\{-1,1\}$ and the augmented data $\{\Tilde{X}_i, \Tilde{Y}_i, \}_{i=1}^{n_1}$ satisfies $(X, Y) \sim N (Y t \mu, \sigma^2 I)$ for $t \in [0, 1]$. We have $n_1 = m \times n_0$. $m \geq 1$, for any $\gamma \geq \frac{2}{\sqrt{m}}[(m - 1) (n_0 / d)^{1/8}]$, we have
% \begin{equation}
%     \mathbb{PE}_a \left(\hat{\theta}_{m_0} \right) \geq \mathbb{PE}_a \left( \Tilde{\theta}_{m_1}\right)
% \end{equation}
% \end{theorem}

%\paragraph{Remark} 
%As we can see, $\gamma$ need to be ...

\section{Additional Experimental Results}
\subsection{Experiments on MNIST dataseet}

\begin{table}[htp]\small
\centering
\caption{Certified accuracy on MNIST dataset.
%\TBD{Directly show the best result, add more variant of ours}
}
\vspace{-5pt}
    \begin{adjustbox}{width=0.99\linewidth}
    \begin{tabular}{cl|cccccccccccc}
    \toprule
    $\sigma$ &  Models (MNIST)  & 0.00 & 0.25 & 0.50 & 0.75 & 1.00 & 1.25 & 1.50 & 1.75 & 2.00 & 2.25 & 2.50 & 2.75 \\ 
    \midrule
    \multirow{9}{*}{0.25} & Gaussian   & 99.2 & 98.5 & 96.7 & 93.3 & 0.0 & 0.0 & 0.0 & 0.0 & 0.0 & 0.0 & 0.0 & 0.0 \\
    & Stability training    & 99.3 & 98.6 & 97.1 & 93.8 & 0.0 & 0.0 & 0.0 & 0.0 & 0.0 & 0.0 & 0.0 & 0.0 \\
    & SmoothAdv  & 99.4 & 99.0 & 98.2 & 96.8 & 0.0 & 0.0 & 0.0 & 0.0 & 0.0 & 0.0 & 0.0 & 0.0  \\
    & MACER   & 99.3 & 98.7 & 97.5 & 94.8 & 0.0 & 0.0 & 0.0 & 0.0 & 0.0 & 0.0 & 0.0 & 0.0 \\ 
    & Consistency  & {99.3} & {98.7} & {98.2} & {95.8} & 0.0 & 0.0 & 0.0 & 0.0 & 0.0 & 0.0 & 0.0 & 0.0 \\
    & {SmoothMix }  & {99.2} & {{98.8}} & {{98.0}} & {96.3} & 0.0 & 0.0 & 0.0 & 0.0 & 0.0 & 0.0 & 0.0 & 0.0 \\
    \cmidrule(l){2-2} \cmidrule(l){3-3} \cmidrule(l){4-14}
    & \textbf{ours}  &99.0	&98.1	&\bf{97.3}	&\bf{95.8}	& 0.0 & 0.0 & 0.0 & 0.0 & 0.0 & 0.0 & 0.0 & 0.0\\
    & \textbf{ours} + SmoothAdv &98.2	&97.1	&96.3	&\bf{94.7}	&0.0	&0.0	&0.0	&0.0	&0.0	&0.0	&0.0	&0.0	\\
    & \textbf{ours} + SmoothMix &98.3	&97.7	&\bf{97.0}	&\bf{96.1}	&0.0	&0.0	&0.0	&0.0	&0.0	&0.0	&0.0	&0.0 \\
    %& \textbf{ours} PCA &98.9 &98.1 &97.2	&95.5 & 0.0 & 0.0 & 0.0 & 0.0 & 0.0 & 0.0 & 0.0 & 0.0\\
    \midrule
    \multirow{9}{*}{0.50} & Gaussian  & 99.2 & 98.3 & 96.8 & 94.3 & 89.7 & 81.9 & 67.3 & 43.6 & 0.0 & 0.0 & 0.0 & 0.0 \\
    & Stability training   & 99.2 & 98.5 & 97.1 & 94.8 & 90.7 & 83.2 & 69.2 & 45.4 & 0.0 & 0.0 & 0.0 & 0.0 \\
    & SmoothAdv & 99.0 & 98.3 & 97.3 & 95.8 & 93.2 & 88.5 & 81.1 & 67.5 & 0.0 & 0.0 & 0.0 & 0.0 \\
    & MACER  & 98.5 & 97.5 & 96.2 & 93.7 & 90.0 & 83.7 & 72.2 & 54.0 & 0.0 & 0.0 & 0.0 & 0.0 \\ 
    & Consistency  & 99.2 & {98.6} & {97.6} & {95.9} & {93.0} & {87.8} & {78.5} & {60.5} & 0.0 & 0.0 & 0.0 & 0.0 \\
    & {SmoothMix } & 98.7 & 98.0 & {97.0} & {95.3} & {92.7} & {{88.5}} & {{81.8}} & {{70.0}} & 0.0 & 0.0 & 0.0 & 0.0 \\
    \cmidrule(l){2-2} \cmidrule(l){3-3} \cmidrule(l){4-14}
    & \textbf{ours}  &98.1	&97.3	&96.2	&\bf{94.8}	&\bf{92.2}	&\bf{87.8}	&\bf{79.5}	&\bf{67.7}	& 0.0 & 0.0 & 0.0 & 0.0   \\
    & \textbf{ours} + SmoothAdv &88.8	&86.7	&84.4	&80.6	&77.6	&73.9	&\bf{70.3}	&\bf{64.0}	&0.0	&0.0	&0.0	&0.0	\\
    & \textbf{ours} + SmoothMix &97.7	&97.0	&95.4	&93.6	&89.1	&\bf{84.9}	&\bf{78.0}	&\bf{67.7} &0.0	&0.0	&0.0	&0.0\\
    %& \textbf{ours} PCA &98.2	&97.2	&95.0	&92.5	&88.9	&83.5	&75.8&	61.4	& 0.0 & 0.0 & 0.0 & 0.0  \\
    \midrule
    \multirow{9}{*}{1.00} & Gaussian  & 96.3 & 94.4 & 91.4 & 86.8 & 79.8 & 70.9 & 59.4 & 46.2 & 32.5 & 19.7 & 10.9 & 5.8 \\
    & Stability training   & 96.5 & 94.6 & 91.6 & 87.2 & 80.7 & 71.7 & 60.5 & 47.0 & 33.4 & 20.6 & 11.2 & 5.9 \\
    & SmoothAdv  & 95.8 & 93.9 & 90.6 & 86.5 & 80.8 & 73.7 & 64.6 & 53.9 & 43.3 & 32.8 & 22.2 & 12.1  \\
    & MACER  & 91.6 & 88.1 & 83.5 & 77.7 & 71.1 & 63.7 & 55.7 & 46.8 & 38.4 & 29.2 & 20.0 & 11.5  \\
    & Consistency  & 95.0 & 93.0 & 89.7 & 85.4 & 79.7 & 72.7 & 63.6 & 53.0 & 41.7 & 30.8 & 20.3 & 10.7 \\
    & {SmoothMix }  & 93.5 & 91.3 & 87.9 & 83.2 & 77.9 & 71.1 & {62.5} & {53.6} & {{44.9}} & {{36.5}} & {{28.8}} & {{21.3}} \\
    \cmidrule(l){2-2} \cmidrule(l){3-3} \cmidrule(l){4-14}
    & \textbf{ours}  &91.7&	88.7&	85.4&	81.1	&75.4	&68.0	&\bf{61.4}	&\bf{52.3}	&\underline{\bf{45.0}}	&\underline{\bf{37.8}}  &	\underline{\bf{30.7}}	&\underline{\bf{23.2}} \\
    & \textbf{ours}+ SmoothAdv &86.2	&82.4	&78.9	&73.9	&67.9	&62.4	&56.8	&\bf{49.4}	&\bf{43.7}	&\underline{\bf{38.4}}	&\underline{\bf{33.0}}	&\underline{\bf{27.8}} \\
    %& \textbf{ours} PCA + SmoothAdv &69.1	&65.1	&59.0	&54.7	&51.2	&47.3	&42.4	&38.4	&34.4	&30.4 &	24.2	&17.5	 \\
    & \textbf{ours} + SmoothMix &92.5	&90.2	&86.5	&83.0	&77.3	&70.6	&\bf{62.6}	&\bf{53.4}	&\underline{\bf{45.9}}	&\underline{\bf{37.8}}	&\underline{\bf{30.7}}	&\underline{\bf{22.5}}\\
    %& \textbf{ours} PCA &90.1	&87.7	&83.6	&77.1	&69.9	&63.6	&54.5	&44.5	&35.3	&27.6	&19.1	&12.8 \\
    \bottomrule
\end{tabular}
    \end{adjustbox}
\label{table:mnist}
\end{table}
%\paragraph{Results on MNIST Dataset} 
In Table~\ref{table:mnist}, we show the comparison results on the MNIST dataset with certified accuracy at various radii, and the comparison results on the CIFAR-10 dataset are shown in Table~\ref{table:cifar10}. 
We set our results \textbf{bold-faced} whenever the value improves the Gaussian baseline and \underline{underlines} whenever the value improves the best among the considered baselines.

As shown in Table~\ref{table:mnist}, our method can significantly improve the certified test accuracy compared with Gaussian \cite{cohen2019certified_cert_rob} on the MNIST dataset, and also outperforms existing methods, i.e., SmoothAdv \citep{salman2019provably}, MACER \citep{Zhai2020MACER}, Consistency \citep{jeong2020consistency}, and SmoothMix \citep{jeong2021smoothmix}. The important characteristic of our method is the robustness under larger noise levels. Our method achieved the highest certified test accuracy among all the noise levels when the radii are large, i.e., radii 0.50-0.75 under noise level $\sigma=0.25$, radii 0.75-1.75 under noise level $\sigma=0.50$, and radii 1.50-2.75 under noise level $\sigma=1.00$, which clearly demonstrated the effectiveness of our data augmentation method, as the robustness improvement under large noise level is more critical~\cite{cohen2019certified_cert_rob}.

We also combined our method with SmoothAdv and SmoothMix to evaluate whether our data augmentation method can provide additive improvements. As shown in Table~\ref{table:mnist}, we find that by combing our data augmentation mechanism, the performance of previous SOTA methods can be even better, which demonstrates the effectiveness of our method, and can be easily used as an add-on mechanism for many other algorithms to improve learning robustness.

\subsection{More ablation studies}

\paragraph{Comparison on training batch size} We show the performance comparison for different training batch sizes on the MNIST dataset.

\begin{table}[H]\small
\centering
\caption{Comparison of different training batch with VAE on the MNIST dataset of certified accuracy at various radii (noise level $\sigma=0.5$).}
\vspace{0.03in}
    \begin{adjustbox}{width=0.9\linewidth}
    \begin{tabular}{cc|ccccccccccc}
    \toprule
      $\sigma$ & Training batch (MNIST)   & 0.00 & 0.25 & 0.50 & 0.75 & 1.00 & 1.25 & 1.50 & 1.75 & 2.00 & 2.25 & 2.50  \\ 
    \midrule
     \multirow{7.5}{*}{0.5} 
    %&16 	\\
    %&32 	\\
    &64 &97.5	&96.5	&95.2	&93.5	&90.4	&84.5	&76.5	&64.4	&0.0	&0.0	&0.0	\\
    &128 &97.7	&96.9	&96.0	&94.5	&91.5	&87.0	&78.7	&65.8	&0.0	&0.0	&0.0	\\
    &256 &97.9	&96.8	&95.9	&94.3	&91.6	&86.5	&79.0	&66.3 &0.0	&0.0	&0.0	\\
    & 512 &97.6	&97.2	&96.1	&94.5	&91.8	&87.0	&78.7	&67.0	&0.0	&0.0	&0.0 \\
    & 1024 &97.9	&97.0	&96.1	&94.1	&91.7	&86.7	&78.4	&66.8	&0.0	&0.0	&0.0 \\
    & 2048 &97.7	&97.2	&96.1	&94.2	&91.8	&86.4	&78.2	&67.0	&0.0	&0.0	&0.0 \\
    & 3072 &97.7	&96.8	&96.0	&94.1	&91.5	&86.2	&78.1	&66.6	&0.0	&0.0	&0.0 \\
    & 4096 &97.4	&96.7	&95.4	&93.5	&91.0	&86.0	&78.1	&66.0	&0.0	&0.0	&0.0 \\
    \bottomrule
\end{tabular}
    \end{adjustbox}
\end{table}

\paragraph{Comparison on augmentation batch size $b_n$} We show the performance comparison for different augmentation batch sizes on the MNIST dataset.

\begin{table}[H]\small
\centering
\caption{Comparison of different augmentation batch size $b_n$ with VAE on the MNIST dataset of certified accuracy at various radii (noise level $\sigma=0.5$).}
\vspace{0.03in}
    \begin{adjustbox}{width=0.99\linewidth}
    \begin{tabular}{cc|ccccccccccc}
    \toprule
      $\sigma$ & augmentation batch $b_n$ (MNIST)   & 0.00 & 0.25 & 0.50 & 0.75 & 1.00 & 1.25 & 1.50 & 1.75 & 2.00 & 2.25 & 2.50  \\ 
    \midrule
     \multirow{8.5}{*}{0.5} 
    &4 &97.9	&96.8	&95.9	&94.4	&91.9	&87.0	&78.7	&67.5	&0.0	&0.0	&0.0\\
    &8 &97.8	&97.2	&95.9	&94.6	&91.6	&87.4	&79.2	&66.7	&0.0	&0.0	&0.0\\
    &16 &97.9	&97.1	&96.1	&94.2	&91.8	&87.8	&78.7	&67.0 &0.0	&0.0	&0.0	\\
    &32 &98.0	&96.9	&95.9	&94.8	&91.7	&87.0	&79.2  &67.1	&0.0	&0.0	&0.0 \\
    &64 &97.8	&96.9	&95.8	&94.2	&92.1	&87.5&	78.6	&67.5	&0.0	&0.0	&0.0	\\
    &128 &97.9	&97.2	&95.9	&94.7	&91.8	&87.5	&78.6	&66.3	&0.0	&0.0	&0.0	\\
    &256 &97.7	&97.1	&96.1	&94.4	&91.9	&88.0	&78.8	&67.6 &0.0	&0.0	&0.0	\\
    & 512 &97.8	&97.2	&96.2	&94.6	&91.8&	87.1	&79.1	&66.8	&0.0	&0.0	&0.0 \\
    & 1024 &98.0	&96.9	&96.2	&94.5	&91.1	&86.7	&78.4	&67.3 	&0.0	&0.0	&0.0 \\
    \bottomrule
\end{tabular}
    \end{adjustbox}
\end{table}

\paragraph{Influence of magnificent coefficient $m_a$} The comparison results of different magnificent coefficient $m_a$  with corresponding certified accuracy at various radii are shown in Table~\ref{table:mag_coe}.

\begin{table}[H]\small
\centering
\caption{Comparison of different magnificent coefficients $m_a$ on the MNIST dataset of certified accuracy at various radii. (noise level $\sigma=0.5$)}
\vspace{0.03in}
    \begin{adjustbox}{width=0.99\linewidth}
    \begin{tabular}{cc|cccccccccccc}
    \toprule
      $\sigma$ & magnificent coefficient $m_a$ (MNIST)  & 0.00 & 0.25 & 0.50 & 0.75 & 1.00 & 1.25 & 1.50 & 1.75 & 2.00 & 2.25 & 2.50  & 2.75\\ 
    \midrule
     \multirow{7}{*}{0.5} 
&1 &97.8	&97.2	&95.9	&94.3	&92.2&87.6	&78.7	&67.5	&0.0	&0.0	&0.0	&0.0\\	
&2 &98.1	&97.3	&96.2	&94.8	&92.2	&87.8	&78.7 &	67.7		&0.0	&0.0	&0.0	&0.0 \\
&3 	&98.2	&97.3	&96.0	&94.7	&92.0	&87.4	&78.6	&66.6	&0.0	&0.0	&0.0	&0.0 \\
&4 	&97.9	&97.1	&95.8	&94.6	&92.2	&87.8	&78.5	&66.8	&0.0	&0.0	&0.0	&0.0 \\
&5 	&98.1	&97.1	&96.1	&94.5	&92.3	&87.4	&79.2	&66.4	&0.0	&0.0	&0.0	&0.0\\
&6 	&97.8	&97.0	&96.1	&94.5	&92.6&	87.5	&79.2	&67.0	&0.0	&0.0	&0.0	&0.0\\
&7 	&98.1	&96.7	&96.0	&94.5	&92.0	&87.7	&78.9	&66.4	&0.0	&0.0	&0.0	&0.0\\
    \bottomrule
\end{tabular}
    \end{adjustbox}
\label{table:mag_coe}
\end{table}

\paragraph{Influence of ot\_lbd} The comparison results of different ot\_lbd with corresponding certified accuracy at various radii are shown in Table~\ref{table:ot_lbd}.

\begin{table}[H]\small
\centering
\caption{Comparison of different ot\_lbd on the MNIST dataset of certified accuracy at various radii. (noise level $\sigma=0.5$)}
\vspace{0.03in}
    \begin{adjustbox}{width=0.99\linewidth}
    \begin{tabular}{cc|cccccccccccc}
    \toprule
      $\sigma$ & ot\_lbd (MNIST)  & 0.00 & 0.25 & 0.50 & 0.75 & 1.00 & 1.25 & 1.50 & 1.75 & 2.00 & 2.25 & 2.50  & 2.75 \\ 
    \midrule
     \multirow{4}{*}{0.5} 
&1e-1 &97.3	&96.1	&94.6	&92.5	&88.3	&82.4	&75.1	&63.0 &0.0	&0.0	&0.0	&0.0 \\	
&1e-2 &96.7	&96.1 & 94.6	&92.3	&88.6	&82.4	&74.6	&64.9 &0.0	&0.0	&0.0	&0.0	\\
&1e-3 &96.8	&96.2	&94.4&	91.6	&88.6 &82.5	&75.1	&63.4	&0.0	&0.0	&0.0	&0.0	\\
&1e-4 &97.2	&96.3	&94.3	&91.3	&87.7	&82.6&	75.4	&64.5	&0.0	&0.0	&0.0	&0.0	\\
    \bottomrule
\end{tabular}
    \end{adjustbox}
\label{table:ot_lbd}
\end{table}

\paragraph{Influence of ot\_t} The comparison results of different ot\_t with corresponding certified accuracy at various radii are shown in Table~\ref{table:ot_t}.

\begin{table}[H]\small
\centering
\caption{Comparison of different ot\_t on the MNIST dataset of certified accuracy at various radii. (noise level $\sigma=0.5$)}
\vspace{0.03in}
    \begin{adjustbox}{width=0.99\linewidth}
    \begin{tabular}{cc|ccccccccccc}
    \toprule
      $\sigma$ & ot\_t (MNIST)  & 0.00 & 0.25 & 0.50 & 0.75 & 1.00 & 1.25 & 1.50 & 1.75 & 2.00 & 2.25 & 2.50  \\ 
    \midrule
     \multirow{5}{*}{0.5} 
&0.1 &97.8	&97.2 &96.1	&94.8 &92.1	&87.6	&78.5	&66.5	&0.0	&0.0	&0.0\\	
&0.2 &97.7	&97.2	&95.9	&94.3	&92.2	&87.5	&78.5	&66.6	&0.0	&0.0	&0.0	\\
&0.3 &97.9	&97.0	&96.2&	94.5	&91.4	&86.5	&78.4	&67.0	&0.0	&0.0	&0.0	\\
&0.4 &97.9	&97.1	&96.1	&93.9	&91.3	&87.3	&78.6	&67.2		&0.0	&0.0	&0.0	\\
&0.5 &98.0	&97.3	&95.9	&93.8	&91.0	&86.7	&77.7	&66.7 &0.0	&0.0	&0.0 \\
    \bottomrule
\end{tabular}
    \end{adjustbox}
\label{table:ot_t}
\end{table}

%\paragraph{Influence of $z$ in VAE} The size of hidden state $z$ in the VAE model also has influence on the performance. We implemented several VAE models and computed the corresponding certified accuracy at various radii at the noise level $\sigma=0.5$, which is shown in Table~\ref{table:vae_z}. More results are shown in the Appendix. 

%\begin{table}[H]\small
%\centering
%\caption{Comparison of different $z$ in VAE on the MNIST dataset of certified accuracy at various radii.}
%\vspace{0.03in}
%    \begin{adjustbox}{width=0.99\linewidth}
%    \input{tables/vae_z}
%    \end{adjustbox}
%\label{table:vae_z}
%\end{table}

\subsection{More experimental results}

\paragraph{Influence of VAE training} The training protocol for VAE can also affect performance. We adopted different VAE training mechanisms and computed the corresponding certified accuracy at various radii, which is shown in Table~\ref{table:vae_train}.

\begin{table}[H]\small
\centering
\caption{Comparison of different VAE training methods on the CIFAR-10 dataset of certified accuracy at various radii.}
\vspace{0.03in}
    \begin{adjustbox}{width=0.99\linewidth}
    \begin{tabular}{cc|cccccccccccc}
    \toprule
      $\sigma$ & VAE training (CIFAR-10)  & 0.00 & 0.25 & 0.50 & 0.75 & 1.00 & 1.25 & 1.50 & 1.75 & 2.00 & 2.25 & 2.50 & 2.75 \\ 
    \midrule
    \multirow{2.5}{*}{0.25} 
     &Without crop and flip   &67.80	&60.70	&51.50	&41.70  &0.00 &0.00 &0.00 &0.00 &0.00 &0.00 &0.00 &0.00\\
     &With crop and flip  &67.20	&60.30	&53.40	&44.60  &0.00 &0.00 &0.00 &0.00 &0.00 &0.00 &0.00 &0.00 \\
     \midrule
     \multirow{2.5}{*}{0.5} 
     &Without crop and flip   &50.70	&46.10	&41.60 &36.50	&32.90	&28.20	&22.40	&18.20 &0.00 &0.00 &0.00 &0.00 \\
     &With crop and flip  &47.50	&44.10	&39.70	&35.50	&30.70	&26.50	&22.40	&18.60  &0.00 &0.00 &0.00 &0.00 \\
     \midrule
     \multirow{2.5}{*}{1.0} 
     &Without crop and flip   &34.90	&31.90	&28.20	&25.20	&23.10	&21.30	&19.20	&17.70	&15.60	&13.60	&11.10	&9.70  \\
     &With crop and flip   &36.40	&33.00	&29.90	&26.90	&24.30	&21.90	&20.00	&17.60&	14.80	&12.90	&10.40	&8.70  \\
    \bottomrule
\end{tabular}
    \end{adjustbox}
\label{table:vae_train}
\end{table}

\paragraph{Influence of the hidden size $z$ in VAE} More ablation study results on the hidden size $z$ in VAE on the  CIFAR-10 dataset are shown in Table~\ref{table:more_z}.

\begin{table}[H]\small
\centering
\caption{Comparison of different hidden sizes $z$ in VAE on the CIFAR-10 dataset of certified accuracy at various radii.}
\vspace{0.03in}
    \begin{adjustbox}{width=0.99\linewidth}
    \begin{tabular}{cc|cccccccccccc}
    \toprule
      $\sigma$ & $z$ in VAE (CIFAR10)  & 0.00 & 0.25 & 0.50 & 0.75 & 1.00 & 1.25 & 1.50 & 1.75 & 2.00 & 2.25 & 2.50 & 2.75 \\ 
    \midrule
     \multirow{4.5}{*}{0.25} 
     &30  &67.90 &	59.40	&48.80	&41.20 &0.00	&0.00	&0.00	&0.00	&0.00	&0.00	&0.00	&0.00\\
     &64  &67.80	&60.70	&51.50	&41.70	&0.00	&0.00	&0.00	&0.00	&0.00	&0.00	&0.00	&0.00  \\
     &90    &67.90	&59.40	&48.90	&41.20	&0.00	&0.00	&0.00	&0.00	&0.00 & 0.00	&0.00	&0.00	 \\
     &128   &69.70	&63.40	&56.00	&47.30	&0.00	&0.00	&0.00	&0.00	&0.00	&0.00	&0.00	&0.00  \\
     \midrule
     \multirow{4.5}{*}{0.5}
     &30  &51.30	&47.00	&41.90	&38.70	&33.80	&28.70	&23.90 &	19.20	&0.00&	0.00	&0.00 &	0.00 \\
     &64   &50.70&	46.10	&41.60	&36.50	&32.90	&28.20&22.40&	18.20	&0.00&	0.00	&0.00 &	0.00 \\
     &90    &51.30	&47.00	&41.90	&38.70	&34.00	&28.60	&23.80	&18.80	&0.00	&0.00	&0.00	&0.00 \\
     &128   &55.40	&50.70	&45.60	&40.60	&35.20	&30.50	&25.70	&19.50	&0.00	&0.00	&0.00	&0.00 \\
     \midrule
     \multirow{4.5}{*}{1.0} 
     &30 &33.80	&31.00	&28.00	&24.60	&22.00	&19.40	&17.30	&14.90	&13.40	&10.40	&9.40	&8.40	  \\
     &64  &34.90	&31.90	&28.20	&25.20&	23.10	&21.30	&19.20	&17.70	&15.60	&13.60	&11.10	&9.70	  \\
     &90   &40.90	&36.90	&34.50	&31.50	&28.10	&24.40	&22.50	&20.40	&16.50	&14.00	&12.30	&9.70 \\
     &128  &39.00	&35.50	&31.90	&28.20	&24.40	&21.80	&18.50	&16.10	&13.80	&11.20	&9.20	&7.70  \\
    \bottomrule
\end{tabular}
    \end{adjustbox}
\label{table:more_z}
\end{table}

\paragraph{Influence of $t$ on MNIST dataset} More ablation study results of different ot\_t with PCA and VAE on the MNIST dataset are shown in the following tables.

\begin{table}[H]\small
\centering
\caption{Comparison of different $t$ with PCA on the MNIST dataset of certified accuracy at various radii (without mixing label, noise level $\sigma=0.25$).}
\vspace{0.03in}
    \begin{adjustbox}{width=0.99\linewidth}
    \begin{tabular}{cc|ccccccccccc}
    \toprule
      $\sigma$ & $t$ (MNIST)  & 0.00 & 0.25 & 0.50 & 0.75 & 1.00 & 1.25 & 1.50 & 1.75 & 2.00 & 2.25 & 2.50  \\ 
    \midrule
     \multirow{10}{*}{0.25} 
     &0.001   &99.40	&98.50	&97.20	&95.00	&0.00	&0.00	&0.00	&0.00	&0.00	&0.00	&0.00  \\
     &0.005  &99.30	&98.30	&96.90	&94.60	&0.00	&0.00	&0.00	&0.00	&0.00	&0.00	&0.00  \\
     &0.10   &99.40	&98.40	&97.20	&95.30	&0.00	&0.00	&0.00	&0.00	&0.00	&0.00	&0.00  \\
     &0.15   &99.50	&98.20	&97.10	&94.50	&0.00	&0.00	&0.00	&0.00	&0.00	&0.00	&0.00  \\
     &0.20   &99.20	&98.70	&97.00	&95.30&0.00	&0.00	&0.00	&0.00	&0.00	&0.00	&0.00	 \\
     &0.25  &99.50	&98.40	&97.00	&94.50	&0.00	&0.00	&0.00	&0.00	&0.00	&0.00	&0.00  \\
     &0.30   &99.30	&98.10	&97.30	&95.10	&0.00	&0.00	&0.00	&0.00	&0.00	&0.00	&0.00 \\
     &0.35  &99.40	&98.60	&96.90	&94.0	&0.00	&0.00	&0.00	&0.00	&0.00	&0.00	&0.00	  \\
     &0.40   &99.60	&98.20	&96.90	&95.00	&0.00	&0.00	&0.00	&0.00	&0.00	&0.00	&0.00 \\
     &0.45   &99.40	&98.30	&97.00	&94.80	&0.00	&0.00	&0.00	&0.00	&0.00	&0.00	&0.00 \\
     &0.50   &99.40	&98.40	&97.20	&95.30	&0.00	&0.00	&0.00	&0.00	&0.00	&0.00	&0.00	 \\
    \bottomrule
\end{tabular}
    \end{adjustbox}
\label{table:more_ot_t_PCA}
\end{table}

\begin{table}[H]\small
\centering
\caption{Comparison of different $t$ with VAE on the MNIST dataset of certified accuracy at various radii (without mixing label, noise level $\sigma=0.25$).}
\vspace{0.03in}
    \begin{adjustbox}{width=0.99\linewidth}
    \begin{tabular}{cc|ccccccccccc}
    \toprule
      $\sigma$ & $t$ (MNIST)  & 0.00 & 0.25 & 0.50 & 0.75 & 1.00 & 1.25 & 1.50 & 1.75 & 2.00 & 2.25 & 2.50  \\ 
    \midrule
     \multirow{10}{*}{0.25} 
     &0.001   &99.30	&98.60	&96.40	&93.30	&0.00	&0.00	&0.00	&0.00	&0.00	&0.00	&0.00  \\
     &0.005  &99.30	&98.20	&97.20	&93.50	&0.00	&0.00	&0.00	&0.00	&0.00	&0.00	&0.00  \\
     &0.10   &98.90	&98.20	&97.00	&93.40	&0.00	&0.00	&0.00	&0.00	&0.00	&0.00	&0.00  \\
     &0.15   &99.10	&98.50	&97.40	&94.00	&0.00	&0.00	&0.00	&0.00	&0.00	&0.00	&0.00  \\
     &0.20   &98.80	&98.50	&97.30	&94.40&0.00	&0.00	&0.00	&0.00	&0.00	&0.00	&0.00	 \\
     &0.25  &99.30	&98.60	&96.80	&94.20	&0.00	&0.00	&0.00	&0.00	&0.00	&0.00	&0.00  \\
     &0.30   &99.10	&98.00	&96.80	&93.80	&0.00	&0.00	&0.00	&0.00	&0.00	&0.00	&0.00 \\
     &0.35  &99.00	&98.40	&97.00	&94.30	&0.00	&0.00	&0.00	&0.00	&0.00	&0.00	&0.00	  \\
     &0.40   &99.20&	98.00	&96.90	&94.50	&0.00	&0.00	&0.00	&0.00	&0.00	&0.00	&0.00 \\
     &0.45   &99.10	&98.20	&96.60	&94.40	&0.00	&0.00	&0.00	&0.00	&0.00	&0.00	&0.00 \\
     &0.50   &99.30	&98.40	&97.00	&94.60	&0.00	&0.00	&0.00	&0.00	&0.00	&0.00	&0.00	 \\
    \bottomrule
\end{tabular}
    \end{adjustbox}
\label{table:more_ot_t_VAE}
\end{table}

\begin{table}[H]\small
\centering
\caption{Comparison of different $t$ with PCA on the MNIST dataset of certified accuracy at various radii (without mixing label, noise level $\sigma=0.5$).}
\vspace{0.03in}
    \begin{adjustbox}{width=0.99\linewidth}
    \begin{tabular}{cc|ccccccccccc}
    \toprule
      $\sigma$ & $t$ (MNIST)   & 0.00 & 0.25 & 0.50 & 0.75 & 1.00 & 1.25 & 1.50 & 1.75 & 2.00 & 2.25 & 2.50  \\ 
    \midrule
     \multirow{10}{*}{0.5} 
     &0.001  &99.00	&98.10	&96.90	&95.40	&91.40	&85.50 & 75.10	&56.20&	0.00	&0.00	&0.00  \\
     &0.005  &98.90	&97.90	&97.10	&95.50	&92.20	&85.70	&75.20	&57.50	&	0.00	&0.00	&0.00	  \\
     &0.10   &99.00	&98.30	&97.20	&95.30  &92.10	&85.60	&75.70	&56.50	&	0.00	&0.00	&0.00	 \\
     &0.15   &99.10	&98.20	&97.10	&95.50	&91.50	&85.30 &74.80 	&55.70	&	0.00	&0.00	&0.00	  \\
     &0.20   &99.00	&98.40 & 96.70	&95.20	&91.90	&86.00	&75.30	&58.20	&	0.00	&0.00	&0.00	 \\
     &0.25  &99.10	&98.10	&97.10	&95.20	&92.10	&85.80	&75.70	&58.00	&	0.00	&0.00	&0.00  \\
     &0.30   &99.10	&98.30	&97.40	&95.30	&91.30	&86.00	&74.60	&57.50	&	0.00	&0.00	&0.00	\\
     &0.35  &99.10	&97.80	&97.10	&95.20	&92.10	&86.30	&75.50	&57.50	&	0.00	&0.00	&0.00	  \\
     &0.40   &99.10	&98.10	&97.00	&95.10	&91.60	&85.90	&75.70	&55.90	&	0.00	&0.00	&0.00	\\
     &0.45   &99.00	&98.30	&97.30	&95.60	&92.50	&85.90	&75.00	&57.30	&	0.00	&0.00	&0.00 \\
     &0.50  &99.00	&98.30&	96.70	&95.40&	91.90&	85.60	&75.00	&57.00	&	0.00	&0.00	&0.00
	 \\
    \bottomrule
\end{tabular}
    \end{adjustbox}
\end{table}

\begin{table}[H]\small
\centering
\caption{Comparison of different $t$ with VAE on the MNIST dataset of certified accuracy at various radii (without mixing label, noise level $\sigma=0.5$).}
\vspace{0.03in}
    \begin{adjustbox}{width=0.99\linewidth}
    \begin{tabular}{cc|ccccccccccc}
    \toprule
      $\sigma$ & $t$ (MNIST)   & 0.00 & 0.25 & 0.50 & 0.75 & 1.00 & 1.25 & 1.50 & 1.75 & 2.00 & 2.25 & 2.50  \\ 
    \midrule
     \multirow{10}{*}{0.5} 
     &0.001  &99.00	&98.100	&96.90	&95.40	&91.40&	85.50	&75.10 &56.20 &	0.00	&0.00	&0.00  \\
     &0.005  &98.90	&97.90&	97.10	&95.50&	92.20	&85.70&	75.20 &57.50	&	0.00	&0.00	&0.00	  \\
     &0.10   &99.00&	98.30&	97.20	&95.30&	92.10&	85.60	&75.70	&56.50	&	0.00	&0.00	&0.00	 \\
     &0.15   &99.10	&98.20&	97.10&	95.50&	91.50	&85.30	&74.80 &	55.70	&	0.00	&0.00	&0.00	  \\
     &0.20   &99.00	&98.40	&96.70	&95.20	&91.90	&86.00&	75.30&	58.20	&	0.00	&0.00	&0.00	 \\
     &0.25  &99.100	&98.10&	97.10&	95.20&	92.10&	85.80	&75.70	&58.00	&	0.00	&0.00	&0.00  \\
     &0.30  & 99.100&	98.30&	97.40	&95.30	&91.30&	86.00&	74.60&	57.50	&	0.00	&0.00	&0.00	\\
     &0.35 & 99.10	&97.80	&97.10&	95.20	&92.10	&86.30&	75.50	&57.50	&	0.00	&0.00	&0.00	  \\
     &0.40   &99.10	&98.10&	97.00	&95.10	&91.60	&85.90	&75.70&	55.90	&	0.00	&0.00	&0.00	\\
     &0.45   &99.00	&98.30	&97.30	&95.60	&92.50	&85.90	&75.00	&57.30	&	0.00	&0.00	&0.00 \\
     &0.50  &99.00	&98.30	&96.70	&95.40	&91.90&	85.60&	75.00 &57.00	&	0.00	&0.00	&0.00
	 \\
    \bottomrule
\end{tabular}
    \end{adjustbox}
\end{table}

\begin{table}[H]\small
\centering
\caption{Comparison of different $t$ with PCA on the MNIST dataset of certified accuracy at various radii (without mixing label, noise level $\sigma=1.0$).}
\vspace{0.03in}
    \begin{adjustbox}{width=0.99\linewidth}
    \begin{tabular}{cc|ccccccccccc}
    \toprule
      $\sigma$ & $t$ (MNIST)   & 0.00 & 0.25 & 0.50 & 0.75 & 1.00 & 1.25 & 1.50 & 1.75 & 2.00 & 2.25 & 2.50  \\ 
    \midrule
     \multirow{10}{*}{1.0} 
     &0.001  &95.90	&93.90& 90.20&	86.40	&80.40&	72.70&	61.70	&50.80	&38.80	&27.50	&16.40	  \\
     &0.005  &96.00	&94.00	&90.20	&86.40	&80.40	&72.80&	61.40&	50.90	&38.40	&28.20	&16.50	  \\
     &0.10   &95.70	&93.80	&90.00	&85.90	&80.10	&72.50	&61.60	&50.90	&39.20	&28.20	&16.90		 \\
     &0.15   &95.60 &93.90	&89.80	&86.20	&80.30	&72.80&	62.10	&50.80	&38.10	&27.20	&15.90		  \\
     &0.20   &96.20	&94.00	&90.40	&86.30	&80.30	&72.40	&61.50	&50.40	&38.40	&27.40	&16.30		 \\
     &0.25  &96.10	&93.90	&90.50	&86.10	&80.20	&72.70	&61.40	&50.80	&38.70	&27.90	&16.00	\\
     &0.30   &96.10	&94.10	&90.00	&86.70	&80.00&	72.40	&61.40	&50.50	&38.10	&27.20	&16.10		\\
     &0.35  &95.90	&93.80	&90.20	&86.40	&80.30	&72.90	&61.40	&50.70	&37.80	&27.600	&15.800		  \\
     &0.40   &96.10	&93.90	&90.20	&86.60	&79.90	&72.50	&61.50	&50.30	&38.10	&28.00	&16.30		\\
     &0.45   &95.80	&93.90	&90.10&	86.60	&80.40	&72.30	&61.30	&50.80	&37.80	&27.40	&16.30	 \\
     &0.50  &96.10	&93.60	&90.00	&86.00	&79.80	&73.20	&61.70	&50.90	&38.30	&27.30	&16.10	 \\
    \bottomrule
\end{tabular}
    \end{adjustbox}
\end{table}

\begin{table}[H]\small
\centering
\caption{Comparison of different $t$ with VAE on the MNIST dataset of certified accuracy at various radii (without mixing label, noise level $\sigma=1.0$).}
\vspace{0.03in}
    \begin{adjustbox}{width=0.99\linewidth}
    \begin{tabular}{cc|ccccccccccc}
    \toprule
      $\sigma$ & $t$ (MNIST)   & 0.00 & 0.25 & 0.50 & 0.75 & 1.00 & 1.25 & 1.50 & 1.75 & 2.00 & 2.25 & 2.50  \\ 
    \midrule
     \multirow{10}{*}{1.0} 
     &0.001  &96.00	&94.20	&90.40	&86.70	&79.70&	73.10&	61.80&	49.90	&35.90	&24.20	&13.30		  \\
     &0.005  &95.70	&93.90	&90.50&	86.20	&80.20	&72.40	&61.80	&49.80	&35.80	&24.90	&13.50	  \\
     &0.10  & 95.60	&93.90&90.00&	86.80	&80.30	&72.60	&61.20	&50.30	&36.00	&24.80	&13.70			 \\
     &0.15   &95.80	&94.00	&90.30&	86.70	&80.30&	72.30	&61.90	&49.90	&6.60	&25.40	&13.70			  \\
     &0.20   &95.50&	94.10	&89.80	&86.20	&79.70	&72.30	&62.00	&50.10	&37.10	&26.100	&14.100		 \\
     &0.25  &95.80&	93.80&	90.10&	86.50&	80.30&	72.70&	62.50&	50.50&	37.10	&25.90	&14.40		\\
     &0.30   &95.80&	93.70&	90.50	&86.00	&80.30&	72.70	&62.10	&50.00	&37.90	&26.60	&14.50			\\
     &0.35  &95.50	&93.50	&89.80	&86.10	&80.40	&72.40	&61.70&	50.50&	37.30	&26.50&	14.60		  \\
     &0.40   &95.80	&93.60&	89.80&	85.60&	79.70	&72.30&	62.10	&51.00	&37.80	&26.80	&14.80		\\
     &0.45  & 95.60	&93.20	&89.70&	86.00	&79.80	&72.20	&62.30	&50.10	&38.40	&28.20	&15.70		 \\
     &0.50  &95.50	&93.30&	89.60&85.80&	79.70&	72.20	&61.80&	51.20	&38.40	&27.80	&16.90		 \\
    \bottomrule
\end{tabular}
    \end{adjustbox}
\end{table}

\end{document}